\documentclass[conference]{IEEEtran}
\usepackage{times}

\usepackage[numbers]{natbib}
\usepackage{multicol}
\usepackage[bookmarks=true]{hyperref}
\usepackage{color}
\let\labelindent\relax
\usepackage{enumitem}
\usepackage{graphicx}
\usepackage{amsmath}
\usepackage{amsfonts}       % blackboard math symbols
\usepackage{
tikz,
relsize,
booktabs
}
\usepackage[font={small}]{caption}
\usepackage{bbm}

\usepackage{algorithm}
\usepackage{algpseudocode}

\usepackage{tikz}

\renewcommand\thefigure{\arabic{figure}}
\newcommand{\xxnote}[3]{}
\ifx\hidenotes\undefined
  \renewcommand{\xxnote}[3]{}
\fi

\hypersetup{
    colorlinks=true,
    linkcolor=blue,
    filecolor=magenta,      
    citecolor=blue,
    urlcolor=teal,
}

\newcommand{\method}{\textsc{FISH}} %Fast Imitation of Skills from Humans
\newcommand{\website}{\href{https://fast-imitation.github.io/}{fast-imitation.github.io}}

\IEEEoverridecommandlockouts

\begin{document}

% paper title
\title{Teach a Robot to FISH: Versatile Imitation from \\ One Minute of Demonstrations}
% FISH: Fast Imitation of robotic Skills from Human demonstrations.

% You will get a Paper-ID when submitting a pdf file to the conference system
% \author{Author Names Omitted for Anonymous Review. Paper-ID [17]}
% \author{Siddhant Haldar\textsuperscript{\textsection}\thanks{Correspondence to: siddhanthaldar@nyu.edu, jp5981@nyu.edu} \qquad Jyothish Pari\textsuperscript{\textsection} \qquad Anant Rai \qquad Lerrel Pinto \vspace{0.1in}
\author{Siddhant Haldar*\thanks{Correspondence to: siddhanthaldar@nyu.edu,  jyo.pari@nyu.edu} \qquad Jyothish Pari* \qquad Anant Rai \qquad Lerrel Pinto \vspace{0.1in}
\\ \vspace{0.1in} * Equal contribution
\\ \vspace{0.1in} New York University
\\{\small \tt \website{}}
\vspace{-0.2in}
}
% \maketitle

\makeatletter
\let\@oldmaketitle\@maketitle%
\renewcommand{\@maketitle}{\@oldmaketitle%
    \centering
    \vspace{0.2in}
    \includegraphics[width=\linewidth]{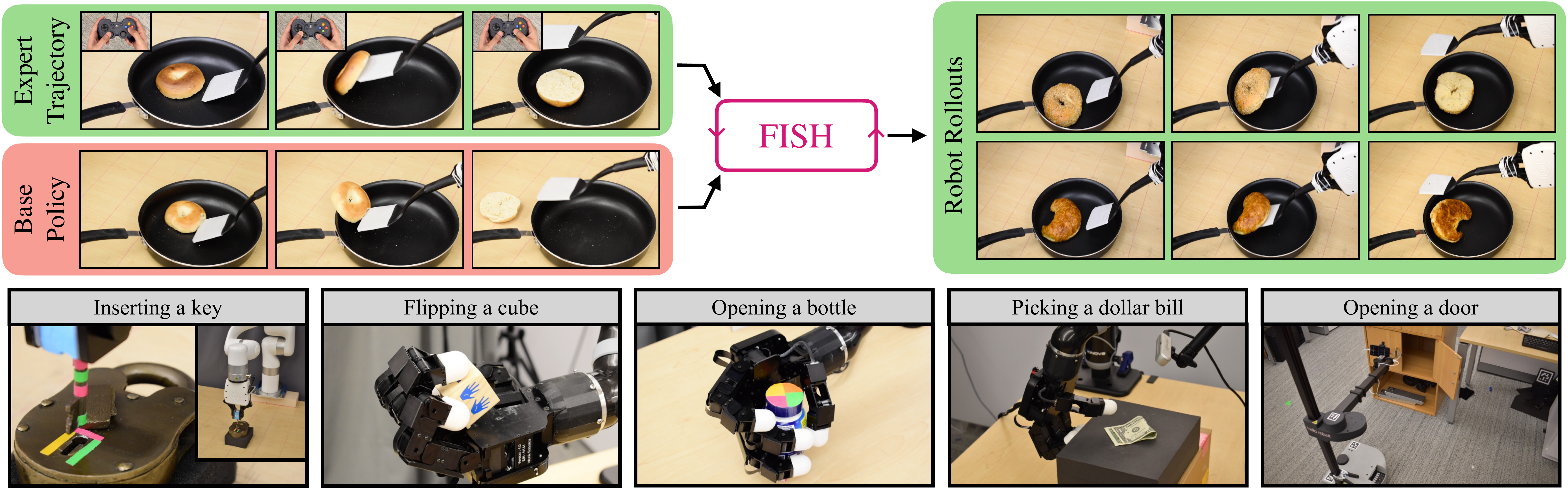}
    \captionof{figure}{Given less than one minute of expert trajectories by a human operator for a task, \method{} can teach skills to a robot that adapt an offline imitated base-policy to new object configurations not seen in the demonstrations. This is done through an interactive RL procedure that improves the visual match between the robot's behavior and the demonstrated trajectory. Both the policy and matching function operate on visual observations. The versatility of \method{} makes it compatible with a variety of robots (e.g. xArm, Allegro, Stretch) and tasks.}
    \label{fig:intro}
}
\makeatother

\maketitle

% \begingroup\renewcommand\thefootnote{\textsection}
% \footnotetext{Equal contribution}
% \endgroup
% \blfootnote{* Equal contribution}

\begin{abstract}
While imitation learning provides us with an efficient toolkit to train robots, learning skills that are robust to environment variations remains a significant challenge. Current approaches address this challenge by relying either on large amounts of demonstrations that span environment variations or on handcrafted reward functions that require state estimates. Both directions are not scalable to fast imitation. In this work, we present Fast Imitation of Skills from Humans~(\method{}), a new imitation learning approach that can learn robust visual skills with less than a minute of human demonstrations. Given a weak base policy trained by offline imitation of demonstrations, \method{} computes rewards that correspond to the ``match'' between the robot's behavior and the demonstrations. These rewards are then used to adaptively update a residual policy that adds on to the base policy. Across all tasks, \method{} requires at most twenty minutes of interactive learning to imitate demonstrations on object configurations that were not seen in the demonstrations. Importantly, \method{} is constructed to be versatile, which allows it to be used across robot morphologies (e.g. xArm, Allegro, Stretch) and camera configurations (e.g. third-person, eye-in-hand). Our experimental evaluations on 9 different tasks show that \method{} achieves an average success rate of 93\%, which is around 3.8$\times$ higher than prior state-of-the-art methods.  
\end{abstract}

\IEEEpeerreviewmaketitle

%===============================================================================
\section{Introduction}
\label{intro}

% Why imitation
Imitation learning has proven to be among the most efficient tools to teach robots complex, dexterous and contact-rich skills. Its applications in robotics already span the fields of manipulation~\cite{johns2021coarse, fang2019survey}, locomotion~\cite{peng2020learning, ratliff2007imitation}, navigation~\cite{wang2019reinforced, karnan2022voila}, and flying~\cite{fan2020learn, schilling2019learning}. Such imitation approaches are now gaining traction in directly learning from high-dimensional visual observations~\cite{haldar2022watch, karnan2022voila, mandi2022towards}. In broad strokes, visual imitation produces a policy that takes an image as input, and outputs actions that control the robot to perform desirable behaviors. Directly reasoning from images allows such methods to be generally applied as they circumvent the need for task-dependent estimation of state or design of features.

% Whats wrong with imitation
But, there is no free lunch. The generality of learning vision-based policies comes at the cost of needing a large number of demonstrations. MIME~\cite{sharma2018multiple} uses 400 demonstrations per task, while robomimic~\cite{mandlekar2021matters} uses 200 demonstrations to train manipulation policies. This scale of data significantly hampers our ability to train multiple skills in reasonable amounts of time. Furthermore, collecting large amounts of demonstrations is physically and cognitively taxing on the human demonstrators due to the nature of available teleoperation frameworks~\cite{arunachalam2022holo}. Hence, getting imitation learning to work with few demonstrations is paramount for practical training of robotic skills.

% Why is imitation data inefficient
To understand why imitation learning requires large amounts of data, let us take a look at one common paradigm -- offline imitation. Methods in this class such as Behavior Cloning (BC)~\cite{pomerleau1998autonomous} or Nearest Neighbor retrieval (NN)~\cite{pari2021surprising} use a supervised learning objective to maximize the likelihood of demonstrated actions given observations in the demonstration. To ensure that the resulting policy is generalizable to varying factors in deployment (e.g. object configurations), the demonstration set used in training will need to span these factors of variation. Without sufficient coverage, which is only possible with large amounts of demonstration data, trained policies often suffer from distribution shift during deployment~\cite{ross2011reduction}.

% How to fix offline imitation
To address the large data requirements of offline imitation and instead imitate with few examples, a promising direction is to adapt policies that were trained offline with online RL~\cite{nair2020awac, haldar2022watch, rajeswaran2017learning}. The hope is that while the offline policy, trained with few demonstrations, would fail in deployment, online RL will allow the policy to improve and adapt to deployment scenarios. But how does the RL algorithm get the rewards needed for adaptation? Constructing a task-specific reward function is one possibility~\cite{rajeswaran2017learning, radosavovic2021state}. However, this strategy may not be applicable in real-world scenarios where states of objects are hard to estimate or reward functions are hard to create.

% Our method
In this work, we present Fast Imitation of Skills from Humans~(\method{}), a new technique for robotic imitation, where given only a minute of demonstrations (between 1 to 3 trajectories), a robot can learn visual policies that both solve the task and adapt to new object configurations through subsequent online training. \method{} operates in two phases. First, a weak base policy is learned by offline imitation on the few demonstrations. Second, a residual policy~\cite{silver2018residual, johannink2019residual, zhang2019deep, alakuijala2021residual} is trained to produce corrective offsets to the weak policy. During online trial and error training, only the residual policy is updated, while the weak policy is queried as a black box model. This allows the use of non-parametric weak policies that are shown to be superior and more robust than parametric ones in low-data settings~\cite{pari2021surprising, arunachalam2022dexterous, arunachalam2022holo}.

An important consideration in online policy learning is obtaining relevant rewards for robot behavior. Since we do not have access to task-specific reward functions, the rewards will need to be inferred from visual data. This is done by matching the visual observations from robot rollouts with the trajectory demonstrated by the human. The matching function uses fast approximations to Optimal Transport (OT)~\cite{cuturi2013sinkhorn} to generate a matching score, which is proportional to the rewards. This procedure does not require explicit estimation of the state or any other object-centric representation.

% Furthermore, the detachment of the base policy from the residual policy prevents any degradation of the base policy from noisy gradients in the beginning of training.\LP{Break this para up to emphasize the OT matching from trajectory of representations.}

We evaluate \method{} on three different robot platforms that cover different morphologies, weak  base policies, camera placements, and gripper types. Through an extensive study across 9 tasks, we present the following key insights:
\begin{enumerate}
    \item \method{} improves upon prior state-of-the-art work in online imitation~\cite{haldar2022watch, cohen2022imitation, kostrikov2018discriminator} with an average of 93\% improvement in success rate given 20 minutes of online interactions (Section~\ref{sec:fish_efficiency}).
    \item We find that \method{} can generalize and adapt to a wide range of object configurations unseen in training (Section~\ref{sec:fish_efficiency}).
    \item Ablations on different representation modules, adaptation strategies, and exploration strategies show that the design decisions in \method{} are crucial for high performance (Section~\ref{experiments}).
\end{enumerate}
Open-sourced code and videos of \method{} can be found at: \\ \website{}.

\section{Background}
\label{background}
 Our work builds on several fundamental ideas in reinforcement learning, imitation learning and optimal transport. Here, we describe the most relevant background for \method{}.

\subsection{Reinforcement Learning (RL)}
We study RL as a discounted infinite-horizon Markov Decision Process (MDP)~\cite{bellman1957markovian, sutton2018reinforcement}. For pixel observations, the agent's observation is approximated as a stack of consecutive RGB frames~\cite{mnih2015human}. The MDP is of the form $(\mathcal{O}, \mathcal{A}, P, R, \gamma, d_{0})$ where $\mathcal{O}$ is the observation space, $\mathcal{A}$ is the action space, $P: \mathcal{O}\times\mathcal{A}\rightarrow\Delta(\mathcal{O})$ is the transition function, $R:\mathcal{O}\times\mathcal{A}\rightarrow\mathbb{R}$ is the reward function, $\gamma$ is the discount factor and $d_{0}$ is the initial state distribution. In this work, we use an actor critic based method to maximize the expected discounted sum of rewards. The rewards obtained through the OT computation can be used to optimize our policy through off-policy learning~\cite{kostrikov2018discriminator}. In this work, we use Deep Deterministic Policy Gradient (DDPG)~\cite{lillicrap2015continuous} as our RL optimizer, which is an actor-critic algorithm that concurrently learns a deterministic policy $\pi_\phi$ and a Q-function $Q_\theta$. Instead of minimizing the one-step Bellman residual as in vanilla DDPG, we use the n-step variant proposed by \citet{yarats2021mastering} which has been successful on visual control problems.

\subsection{Imitation Learning~(IL)}
In imitation learning, the goal is a learn a behavior policy $\pi^b$ from either an expert policy $\pi^e$ or trajectories derived from an expert policy $\mathcal{T}^e$. In this work, we operate in a setting where the agent only has access to expert observational trajectories, i.e. $\mathcal{T}^e \equiv \{(o_t, a_t)_{t=0}^{T}\}_{n=0}^N$. Here, N refers to the number of trajectory rollouts and T denotes the episode length. We opt for this specific setting since obtaining expert or near-expert demonstrations is feasible in real-world settings~\cite{zhan2020framework,young2020visual} and is in line with recent works in the area~\cite{haldar2022watch,dadashi2020primal,ho2016generative,kostrikov2018discriminator}.

\subsection{Inverse Reinforcement Learning~(IRL)} 
IRL~\cite{ng2000algorithms,abbeel2004apprenticeship} reformulates the IL problem in the RL setting by inferring the reward function $r^e$ from expert trajectories $\mathcal{T}^e$. The inferred reward $r^e$ is used to derive the behavior policy $\pi^b$ using policy optimization. Prominent algorithms in IRL~\cite{kostrikov2018discriminator,ho2016generative} require alternating the inference of reward and optimization of policy in an iterative manner, which is practical for restricted model classes~\cite{abbeel2004apprenticeship}. For compatibility with more expressive deep networks, techniques such as adversarial learning~\cite{ho2016generative,kostrikov2018discriminator} or optimal-transport~\cite{papagiannis2020imitation,dadashi2020primal,cohen2022imitation} are needed. Adversarial IRL approaches infer a reward by learning a discriminator that minimizes the gap between expert trajectories $\mathcal{T}^e$ and behavior trajectories $\mathcal{T}^b$. Such a learning procedure results in non-stationary rewards $r^e$ for the optimization of $\pi^b$ which is prone to unstable training.

\subsection{Optimal Transport (OT) for imitation}
\label{subsec:OT}
In order to alleviate the non-stationary reward issue with adversarial IRL frameworks, we resort to optimal transport (OT) based reward inference in this work~\cite{cuturi2013sinkhorn}. A detailed description of optimal transport is provided in Appendix~\ref{appendix:optimal}. OT seeks to find a way to transform one distribution into another for a given cost function. The cost function represents the cost of transporting mass from one location to another. In our work, we use OT to compute a similarity between an expert trajectory $\mathcal{T}^e = \{o_1^e,...,o_n^e\}$ and a rollout trajectory $\mathcal{T}^b = \{o_1^b,...,o_n^b\}$ from our policy. Each visual observation $o_i^j$ is passed through an encoder to obtain a lower dimensional representation  $z_i^j$. The cost function is computed as a cosine distance between the encoded representations of the observations from two trajectories, and the cost matrix $C$ comprises the costs for different pairs of representations. 

Optimal transport computes a transport plan $\mu^*$ that finds the best matching between $\mathcal{T}^e$ and $\mathcal{T}^b$, where $\mu^*_{i,j}$ represents the strength of the match between the $i^{\text{th}}$ representation from the expert trajectory and the $j^{\text{th}}$ representation from the rollout trajectory under some constraints which are described in Appendix~\ref{appendix:optimal}. We compute rewards from $\mu^*$, by the following equation. 

\begin{equation}
\begin{aligned}
    r^{\text{OT}}(\mathcal{T}^b) = \sum_{t,t'=1}^{T}C_{t,t^{'}}\mu^*_{t,t'}
\end{aligned}
\label{eq:reward}
\end{equation}

Intuitively, maximizing this reward incentivizes the imitation agent to produce trajectories that are closer to the demonstrated trajectories.

\section{Approach}
\label{approach}

%
% \LP{TODO: Introduce the overall framework, i.e. what is given to us. Then outline the approach sketch. Then go into the details. Create Algorithm. Create subsections for Phase 1 (done), Phase 2 (done), Stabilizing OT with Representation Learning, Guided Exploration (maybe?), anything else important in our method that we are missing? }

Given a few demonstrations for complex, contact-rich manipulation that covers a small subset of possible object configurations, we seek to learn a robot policy that can generalize to a larger set of configurations not seen during the demonstrations. To enable this, we propose Fast Imitation of Skills from Humans~(\method{}). \method{} operates in two phases. In the first phase, a weak base policy is trained on the few demonstrations using supervised learning. This weak policy, while being poor in generalization, serves as a useful prior for subsequent adaptation. In the second phase, a residual policy is trained to adapt the base policy to new object configurations. This is done by RL on the robot with these configurations using visual trajectory matching scores as the reward signal. 

% A significant challenge in performing complex, contact-rich manipulation is needing fine control. For instance, consider the task of opening a lock. It is fairly simple to reach the keyhole with sufficient proximity or to open the lock once the key is inside the hole. However, getting the key inside the hole is what poses the primary challenge. Deriving inspiration from this, we propose Fast Imitation of Skills from Humans or \method{}, which uses an imperfect base policy to perform the coarse portions of the task while learning a residual policy to apply fine corrections to this coarse trajectory. This is done in multiple phases. In the first phase, this imperfect base policy is obtained through imitation learning on a single minute of expert demonstrated data. In the second phase, the residual policy is trained on top of this fixed base policy using an IRL objective. The algorithm has been described in Algorithm~\ref{alg:fish} and details of each component along with additional adjustments to stabilize training have been described below.

\begin{figure}
  \begin{center}
    \includegraphics[width = \linewidth]{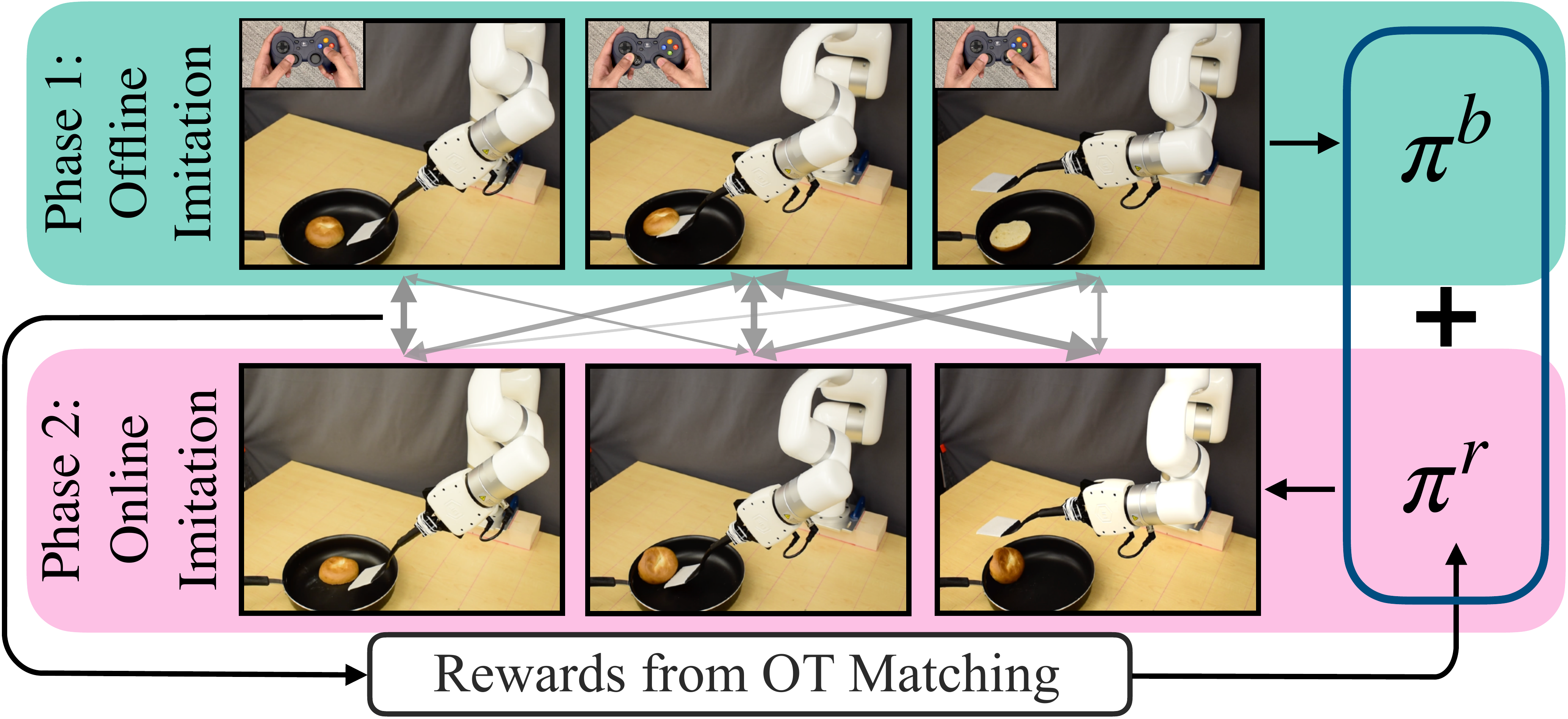}
  \end{center}
  \caption{A schematic of \method{}. The first phase obtains a base policy through offline imitation from demonstrations. The second phase learns a residual model from online interactions}
\label{fig:schematic}
\end{figure}

\begin{figure*}[th]
  \begin{center}
    \includegraphics[width = \linewidth]{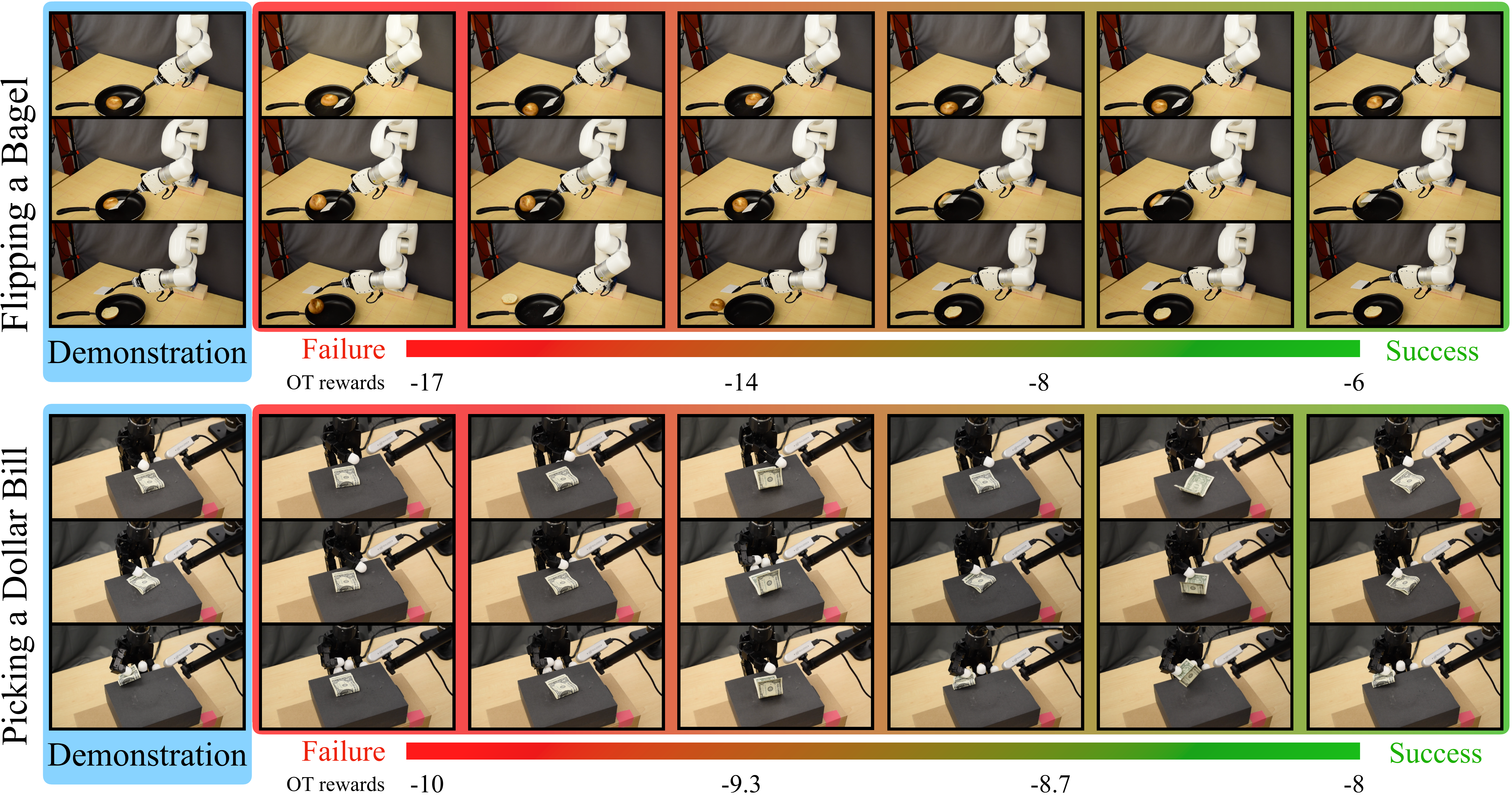}
  \end{center}
  \caption{An analysis of the values of OT rewards for different trajectories with respect to a given expert demonstration. The leftmost column depicts the visual demonstration, while the other columns each depict a trajectory rollout. Trajectories are sorted in increasing order of OT rewards from left to right. Raw OT scores can be visualized using the red-to-green color map.}
\label{figure:ot_reward}
\end{figure*}

\subsection{Phase 1: Non-parametric base policy}
The expert demonstrations are first used to derive an imperfect base policy $\pi^b$. In this work, we stick to non-parametric base policies owing to their proven robustness in the low-data regime~\cite{pari2021surprising, arunachalam2022dexterous, arunachalam2022holo} as compared to parametric alternatives such as Behavior Cloning (BC). We observe that different base policies perform differently across robots and thus, we employ two variants of non-parametric base policies in this work - an open-loop policy and closed-loop Visual Imitation through Nearest Neighbors (VINN)~\cite{pari2021surprising}. More details about these base policies have been provided in Section~\ref{subsec:choice_of_base_policy}.

% We employ three variants of non-parametric base policies in this work.
% \LP{Need to remove this enumeration}
% \begin{enumerate}
%     \item \textbf{Open-loop:} The openloop policy copies the actions performed by the expert at each step of the trajectory.
    
%     \item \textbf{Closed-loop VINN:} Closed-loop VINN or Visual Imitation through Nearest Neighbors(VINN)~\cite{pari2021surprising} converts each visual observation in the demonstration into an encoded representation. During rollouts, $k$-Nearest Neighbor ($k$NN) is used to match to the $k$ closest observations and the action is computed using Locally Weighted Regression (LWR)~\cite{atkeson1997locally} on the actions of the matches observations.
    
%     \item \textbf{Open-loop VINN:} In open-loop VINN, only the first observation of the current trajectory is matched to the closest demonstrated observation. For the remaining steps, the actions from the trajectory corresponding to the matched observation are copied.
% \end{enumerate}

\textbf{Visual representation learning:} Since we operate in the visual domain, a BC policy is trained on the expert demonstrations and we use the encoder from the BC policy to encode the visual observations $o$ into lower dimensional representations $z$. The encoded representation $z$ is provided as an input to both the base policy $\pi^b$ and the residual policy $\pi^r$. An ablation study comparing the use of such a BC encoder with other self-supervised learning techniques~\cite{grill2020bootstrap} as well as pretrained encoders~\cite{deng2009imagenet, Xiao2022, Radosavovic2022, nair2022r3m} is provided in Section~\ref{experiments}.

% Using collected demonstrations, we use non-parametric methods as our base policy. Non-Parametric policies have been showed to be robust on task like dexterous manipulation \cite{} and benchmarking works \cite{}. The models we use are VINN \cite{} and an open loop policy that replicates the trajectory from the demonstration. The open loop method is only applicable in the settings where we have one demonstration. We denote the base policy as $\pi_\text{base}$. 

\subsection{Phase 2: Online offset learning with IRL}
Given the base policy $\pi^b$, we then train a residual policy $\pi^r$ on top of the base policy through environment rollouts. Since we are operating without explicit task rewards, we obtain rewards using OT-based trajectory matching, as described in Section~\ref{background}. A standard RL optimizer utilizes these OT-based rewards $r^{OT}$ to optimize the residual policy $\pi^r$ by maximizing the cumulative reward from the final policy $\pi^{\method{}}$. Similar to prior work~\cite{haldar2022watch, cohen2022imitation}, we use n-step DDPG~\cite{lillicrap2015continuous} as our RL optimizer, a deterministic actor-critic based method that provides high performance in continuous control~\cite{yarats2021mastering}.

\textbf{Residual learning:}
In residual RL~\cite{silver2018residual, johannink2019residual, zhang2019deep, alakuijala2021residual}, given a base policy $\pi^b: \mathcal{Z}\rightarrow\mathcal{A}$ with encoded representations $z\in\mathcal{Z}$ and action $a\in\mathcal{A}$, we learn a residual policy $\pi^r: \mathcal{Z}\times\mathcal{A}\rightarrow\mathcal{A}$ such that an action sampled from the final policy $\pi$ is the sum of the base action $a^b\sim\pi^{b}(z)$ and the residual offset $a^r\sim\pi^{r}(z, a^b)$. In prior work, the base policy $\pi^b$ is either a hand-crafted controller~\cite{silver2018residual, johannink2019residual} or a learned policy~\cite{alakuijala2021residual}. In this work, we use the non-parametric base policy $\pi^{b}$ and learn a residual policy $\pi^{r}$ using OT rewards to refine the action output by $\pi^{b}$. 

\textbf{OT-based reward maximization:} For the RL algorithm, the rewards corresponding to an agent trajectory are computed using the OT-based approach described in Equation~\ref{eq:reward}. A visualization of OT rewards has been shown in Figure~\ref{figure:ot_reward}. These rewards are used to optimize the residual policy using the learning objective shown in Equation~\ref{eq:residual}.

\begin{equation}
    \pi^{r} = \operatorname*{argmax}_\pi \mathbb{E}_{(z,a^b,a^r)\sim \mathcal{D}_{\beta}}[Q(z,a^b,a^r)]
    \label{eq:residual}
\end{equation}

Here, $Q(z,a^b,a^r)$ represents the Q-value from the critic used in actor-critic policy optimization. For an encoded representation $z$ corresponding to observation $o$, $a^b\sim\pi^{b}(z)$ is the action from the base policy, and $a^r\sim\pi^{r}(z,a^b)$ is the action from the residual policy. The executed action $a$ is a sum of $a^b$ and $a^r$.

\textbf{Stabilizing OT with representation learning:} The OT rewards used for the IRL optimization are computed using the encoded representations. As a result, a changing encoder during training results in non-stationary rewards which makes the training prone to instabilities. In order to alleviate this issue, we fix the BC encoder obtained from the demonstrations and the OT rewards are computed using the representations from this fixed encoder. Section~\ref{subsec:insights} shows that a fixed encoder improves stability resulting in superior performance. 

\textbf{Guided exploration for residual policy:}
In contrast to fine-tuning a base policy~\cite{haldar2022watch}, applying offsets through a residual policy allows us to guide the exploration during online learning by injecting domain knowledge into the framework. For instance, if there is only a subspace of the full action space that we need to explore, our framework allows learning only the offsets for this subspace while keeping the base action along the remainder of the action space unaltered. We have provided ablation studies in Section~\ref{subsec:controlled_exp} showing the advantage of such guided exploration. In addition to performance gains, constraining the offsets prevents the robot from going into undesirable positions and enables safer exploration during online learning (refer to Appendix~\ref{appendix:sec:safe_exploration}).

\begin{figure*}[ht!]
  \begin{center}
    \includegraphics[width = \linewidth]{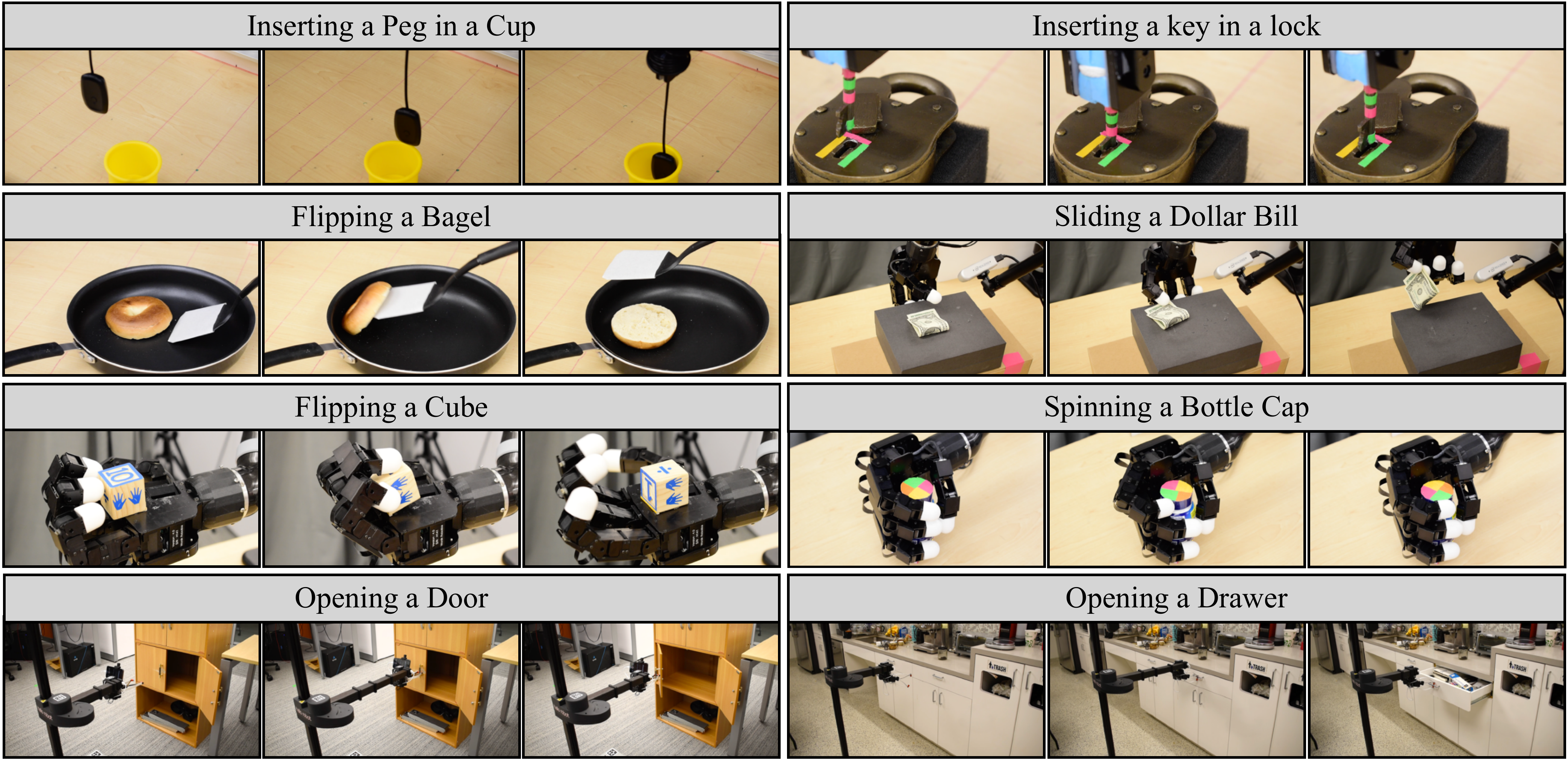}
  \end{center}
  \caption{A visualization of rollouts from \method{} on a selected set of 8 tasks.}
\label{figure:tasks}
\end{figure*}

\section{Experiments}
\label{experiments}

\begin{figure*}[ht!]
  \begin{center}
    \includegraphics[width = \linewidth]{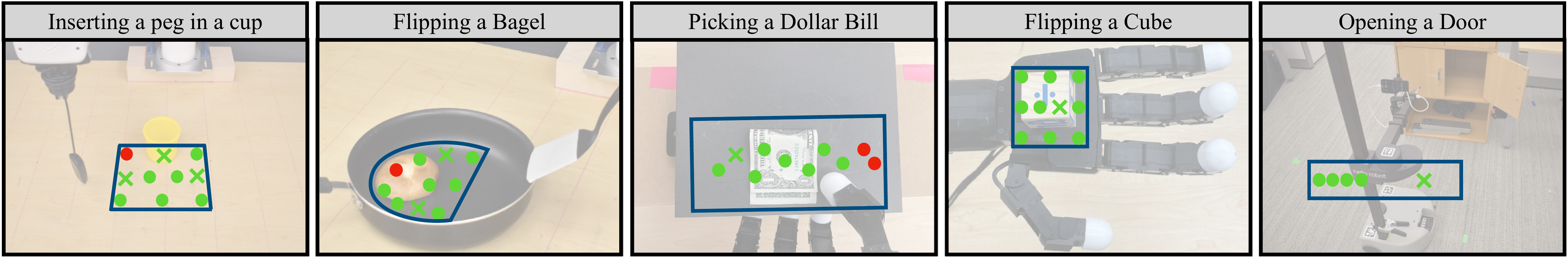}
  \end{center}
  \caption{Plot showing variation in object positions or robot initializations for selected tasks. The region of operation for each task is denoted by the blue box. $\times$ on the images indicate positions where the demonstrations are collected. The green marks indicate positions where \method{} succeeds and the red ones indicate failure modes. As shown, \method{} succeeds with varied object positions and initial robot configurations. }
\label{fig:exploration}
\end{figure*}

% \LP{TODO: Add placeholder tables and figures. Detailed experiment setup. Figures for : Rollouts of our model across tasks. Rollouts for our model vs. strong baselines. Some qualitative understanding of visual matching, Generalization vs time. Examples of generalization (and failures) to new objects (different notes, different types of bagels / bread / food). Tables/bar plots for: cumulative results across tasks, ablations}

% \begin{figure*}[ht!]
%   \begin{center}
%     \includegraphics[width = \linewidth]{figures/exploration.pdf}
%   \end{center}
%   \caption{Plot showing variation in object positions or robot initializations for selected tasks. We show the area of possible locations within the blue rectangle. In the first row, the green dot represents the location where the demonstration was collected. Red dots represent points where our method failed. As shown, our method succeeds on a variety of different object positions or robot starting positions. }
% \label{fig:exploration}
% \end{figure*}

Our experiments are designed to answer the following questions in detail: (1) How efficient is \method{} for imitation learning? (2) How important is guided exploration for faster convergence? (3) How does the choice of base policy affect performance? (4) Are off-the-shelf pretrained encoders useful for online learning in a low-data regime? (5) How do additional implementation details affect \method{}? (6) Does FISH generalize to new objects?

% To understand the capabilities of \method{} and its ability to imitate from unseen object scenarios, we designed a set of experiments to gain insight on these fronts. Specifically, we use three robots and peform 8-9 different tasks of wide variety to demonstrate the versatility of our method on these tasks. In addition, we perform ablation experiments to shed light on the importance of the various components in our method. Specially, our experiments try to answer the following questions. $(a)$ How efficient is \method{} for imitation learning?
% $(b)$ How effective and efficient are learning offsets as compared to finetuning base policy? $(c)$ Is performing regularization on the actor \cite{haldar2022watch} beneficial for learning offsets? $(d)$ Are SSL pretrained encoders on external datasets beneficial for imitation learning$(e)$ Is controlled exploration more efficient? 

\subsection{Experimental setup} We demonstrate the versatility of our algorithm by evaluating our approach on a suite of 9 tasks of varying difficulty across three different robot morphologies. We collect 1 minute of demonstrations (between 1 to 3 trajectories) for each task and allow a maximum of 20 minutes of online learning. For all tasks, we operate purely in the visual domain. 

% The methodology used for demonstration is different across different robots and tasks and more details about them has been provided in Section~\ref{subsec:robot_setup}.

% \subsection{Experimental Setup} We use three robots which are, an Allegro Hand, xArm, Hello robot \cite{}, which have significantly different morphologies. In addition, due to the varying morphologies, the nature of the tasks performed on each robot are significantly different as well. Consequently, we employ different methods to collect demonstrations, such as \cite{arunachalam2022holo} and \cite{young2020visual}, and our demonstrations were collected in less than one minute. Our demonstrations are purely visual, and we do no use any proprioceptive information. In addition, since we have a spectrum of complexity in ours tasks we use from one to three demos.

% We demonstrate the efficacy of \method{} in generalizing to varied object configurations unseen in the expert demonstrations. In Figure~\ref{fig:variations}, we show this generalization for a representative task on the xArm and the Allegro Hand. Even though the expert demonstrations only sparsely cover the operating region, we observe to \method{} generalizes to a majority of the designated area of operation.

\subsection{Robot setup and task descriptions} 
\label{subsec:robot_setup}
We evaluate our approach on 3 different robots - a Ufactory xArm 7 robot, an Allegro Hand, and a Hello Robot Stretch. 

\begin{enumerate}[label=(\alph*),leftmargin=*]
    \item \textbf{Ufactory xArm 7:} We use a xArm 7 robot with a two-fingered gripper for three tasks - key insertion, flipping a bagel, and peg in a cup. The observations are RGB images from a fixed external camera. For each task, the start position of the xArm is fixed and the object position is varied across trajectories. We use closed-loop VINN~\cite{pari2021surprising} as a base policy on the xArm. We provide one, two, and three expert demonstrations for the task of inserting a key, flipping a bagel, and inserting a peg in a cup respectively.
    % The number of expert demonstrations used for inserting a key, flipping a bagel, and inserting a peg in a cup is 1, 2, and 3 respectively. 
    
    % As aforementioned we evaluate our method on three different robots. On the xArm, we perform peg insertion, bagel flipping, and key insertion.In each task we vary the objects position. These tasks are designed to evaluate our model's ability to succeed on, large variations of object positions, dynamic behaviors, and high precision control. On these tasks we find that closed loop VINN \cite{pari2021surprising} is the best base model. 
    
    \item \textbf{Allegro Hand:} We use a 4-fingered robotic hand with a 16-dimensional joint space. We study 3 dexterous manipulation tasks on the hand - cube flipping, bottle cap spinning, and dollar bill picking. The tasks have been designed to exhibit the need for dexterity to accurately manipulate the objects. The observations are RGB images from a fixed external camera. For each task, the start position of the hand is fixed and the object position is varied across trajectories during online training. We use an open-loop policy as a base policy on the hand and the demonstrations are collected using a virtual reality (VR) framework~\cite{arunachalam2022holo}. We use one expert demonstration for all tasks on the Allegro hand.
    
    % On the Allegro hand we design tasks around dexterous manipulation. We use tasks similar to those from \cite{arunachalam2022holo}. Ours tasks are cube flipping, bottle cap spinning, and dollar bill picking. These tasks require dexterity to accurately manipulate the objects. Through the dexterous manipulation tasks, we demonstrate our model's ability to, improve a suboptimal base policy, learn representations on complex visual observations, and perform precise dexterous manipulation with complex interactions. To collect demonstrations we use the framework from \cite{arunachalam2022holo} which utilize a VR headset to allow humans to teleop the hand. In addition, we only use one demonstration for all our tasks, and as our base policy with run an openloop trajectory that copies the expert trajectory. 
    
    \item \textbf{Hello Robot Stretch:} We use Hello Robot's Stretch to showcase our model's ability to interact with a realistic environment using a non-stationary robot. We perform three tasks using the Stretch Robot - door opening, drawer opening, and light switching. The observations are RGB images from an egocentric camera attached to the robot gripper. Hence, the camera viewpoint changes as the robot moves. For each task, the robot is initialized at a random position in front of the object. The demonstration we collect has the robot centered with respect to the handle of the door and drawer and the light switch. We used an open-loop policy as a base policy. We use one expert demonstration for all tasks on the Stretch robot.
    
    % \paragraph{Hello Robot} We use the Hello Robot to showcase our model's ability to interact with realistic home environments. We use only one demonstration as well on the two tasks, which are door opening and drawer opening. The demonstration we collect has the robot centered with respect to the handle of the door and drawer. During training we change the robot's starting position.In addition, we use both an openloop base policy that copies the expert trajectory as well as a version of VINN that only queires the the most similar observation from the demonstration once and copies the rest of the demonstration from that observation.
\end{enumerate}

For each task, we vary the position of the object or the robot at the start of each episode of online learning. All the methods are evaluated on the same initial object or robot configurations, shown in Figure~\ref{fig:exploration}.

% \begin{figure*}[ht!]
%   \begin{center}
%     \includegraphics[width = \linewidth]{figures/task_figure.pdf}
%   \end{center}
%   \caption{A visualization of rollouts from \method{} on a selected set of 8 tasks.}
% \label{figure:tasks}
% \end{figure*}

\begin{table*}[!bhtp]
\caption{\label{table:base_experiments} We report the success rate on each task across different methods. The reported results are based on 5 evaluation trials on the Stretch and 10 evaluation trials each on the Allegro Hand and xArm. Candidate algorithms for base policies have been marked with [B].}

\begin{center}
\begin{tabular}{c|ccc|ccc|ccc}
\hline
                      & \multicolumn{3}{c|}{Stretch Robot}                                                                                                                                                & \multicolumn{3}{c|}{Allegro Hand}                                                                                                                                                         & \multicolumn{3}{c}{xArm}                                                                                                                                                       \\ \hline
Method                & \begin{tabular}[c]{@{}c@{}}Door \\ Opening\end{tabular} & \begin{tabular}[c]{@{}c@{}}Drawer \\ Opening\end{tabular} & \begin{tabular}[c]{@{}c@{}}Light\\ Switching\end{tabular} & \begin{tabular}[c]{@{}c@{}}Cube \\ Flipping\end{tabular} & \begin{tabular}[c]{@{}c@{}}Bottle Cap\\  Spinning\end{tabular} & \begin{tabular}[c]{@{}c@{}}Dollar Bill\\ Picking\end{tabular} & \begin{tabular}[c]{@{}c@{}}Peg in\\ a Cup\end{tabular} & \begin{tabular}[c]{@{}c@{}}Bagel \\ Flipping\end{tabular} & \begin{tabular}[c]{@{}c@{}}Key\\ Insertion\end{tabular} \\ \hline
Open-loop {[}B{]}      & 0.2 & 0.2 & 0.2 & 0.1 & 0.0 & 0.2 & 0.1 & 0.1 & 0.3 \\
VINN BC {[}B{]}       & 0.2 & 0.2 & 0.2 & 0.1 & 0.0 & 0.1 & 0.3 & 0.3 & 0.3 \\
VINN ImageNet {[}B{]} & 0.2 & 0.0 & 0.2 & 0.1 & 0.0 & 0.1 & 0.3 & 0.0 & 0.3 \\
BC {[}B{]}            & 0.2 & 0.0 & 0.0 & 0.0 & 0.0 & 0.0 & 0.5 & 0.3 & 0.3 \\
ROT                   & 0.0 & 0.0 & 0.6 & 0.0 & 0.0 & 0.0 & 0.5 & 0.5 & 0.6 \\
RDAC                  & 0.0 & 0.0 & 0.0 & 0.0 & 0.0 & 0.0 & 0.4 & 0.0 & 0.0 \\
FISH (Ours)           & \textbf{1.0} & \textbf{1.0} & \textbf{1.0} & \textbf{1.0} & \textbf{1.0} & \textbf{0.8} & \textbf{0.9} & \textbf{0.9} & \textbf{0.8}     \\\hline                                      
\end{tabular}
\end{center}
\end{table*}

\subsection{Baseline algorithms}

% \textbf{Base Policies}
We now describe the various imitation learning algorithms, both offline and online, used in this work.
\begin{enumerate}[label=(\alph*),leftmargin=*]
    \item \textbf{Open-loop:} In settings where we have one demonstration, an open-loop policy copies the actions performed by the expert at each step of the trajectory. Though this yields robust performance when the object and robot's positions match the demonstration, it yields poor performance on any variations of the task. 

    \item \textbf{Behavior Cloning (BC):} This refers to the behavior-cloned policy~\cite{pomerleau1998autonomous} trained on expert demonstrations.
    
    % \item \textit{Open Loop VINN:} At the first time step, we query the most similar representation of the observation from the demonstration and copy the rest of the demonstration. The purpose of this method is to allow for the capability to copy a partial trajectory, which allows for more versatility. 

    \item \textbf{Closed-loop VINN:} In closed-loop VINN~\cite{pari2021surprising}, each visual observation in the demonstration is encoded into a representation. During rollouts, the $k$-Nearest Neighbors ($k$NN) algorithm is used to match to the $k$ closest observations, and the action is computed using Locally Weighted Regression (LWR)~\cite{atkeson1997locally} on the actions of the matched observations. In this work, we use a BC encoder for obtaining visual representations.

  \item \textbf{ROT:} ROT~\cite{haldar2022watch} is an IRL algorithm that finetunes a BC pretrained policy through online learning in an environment by leveraging optimal transport for reward computation. ROT gets around the ``forgetting problem'' in such a finetuning setting~\cite{nair2020awac, uchendu2022jump} by using a soft Q-filtering based approach to prevent the actor from incorrectly deviating from the expert demonstration. 

  \item \textbf{RDAC:} Discriminator Actor Critic (DAC)~\cite{kostrikov2018discriminator} is an adversarial imitation learning method ~\cite{ho2016generative,torabi2018generative,kostrikov2018discriminator}. DAC outperforms prior work such as GAIL~\cite{ho2016generative} and AIRL~\cite{fu2017learning}. RDAC is a DAC with a ROT-like regularization applied to it and has been observed to be a strong adversarial IRL baseline~\cite{haldar2022watch}.
  
  % We use a variety of base policies depending on which robot is used. The base policies we use are, Open Loop, Open Loop VINN, and VINN BC Encoder. We then learn a offset model which is trained through IRL which corrects the base policy actions with a residual. Our model uses the encoder trained from BC, and we use regularized optimal transport to compute trajectory similarity as our reward.  In each task we need to set bounds on the offsets as well as choose which subspace of the action space we want the offsets to exist on. This allows for safer and controlled exploration. 
  
\end{enumerate}

\begin{table}[t!]
\caption{\label{table:bc_reg} Comparison between success rates for 10 trials on two tasks for \method{} with and without adaptive regularization of the offsets.}
\centering
\begin{tabular}{ccc}
\hline
Offset Regularization & Bagel Flipping & Dollar Bill Picking \\ \hline
\checkmark & 0.4            & 0.4                 \\
 $\times$ & 0.9            & 0.8                 \\ \hline
\end{tabular}
\end{table}

\subsection{How efficient is \method{} for imitation learning?}
\label{sec:fish_efficiency}
Performance of \method{} on a suite of 9 real-world tasks across 3 different robots has been depicted in Table~\ref{table:base_experiments}. We observe that \method{} outperforms prior work on all tasks. \method{} significantly outperforms ROT~\cite{haldar2022watch}, which is a method for finetuning a pretrained BC policy using online learning. This highlights the benefits of fixing a base policy as compared to modifying it during online finetuning. Further, aligned with results in ~\citet{arunachalam2022dexterous}, we observe that BC performs poorly on the Allegro Hand owing to its high dimensional action space and limited demonstrations. This provides a case for using non-parametric base policies as opposed to parametric alternatives in such low-data regimes. Poor BC performance also affects online learning shown by the poor performance of ROT and is further shown in Section~\ref{subsec:choice_of_base_policy}. We observe that while the learned BC policy is not robust enough to perform with high precision, the resulting representations are still sufficient for downstream fine-tuning (indicated by OT rewards shown in Figure~\ref{figure:ot_reward}). Empirically, we notice that BC is able to complete the coarse portions of the task such as reaching the object. However, the actions are often inaccurate indicating that the BC policy learned on top of the encoded representations is not precise enough.

% We observe that even though BC by itself performs poorly, the representations it learns provide a sufficient signal for task completion as shown in Figure \ref{figure:ot_reward}. 

% In addition, while BC as a base policy performs poorly, as shown in Table \ref{table:base_experiments}, the representations it learns provide a sufficient signal of task completion as shown in Figure \ref{figure:ot_reward}. We notice that BC completes the coarse portions of the task like going towards the object, but for precise actions it often is inaccurate. Therefore, the representations it learns  may not be detailed enough to perform well in moments of high precision, yet the representations still sufficient for our model. 

% As shown in \ref{table:base_experiments} we show that our model is the most competitive across tasks. On the Allegro Hand tasks, Behavior Cloning does not perform well due to the high dimension action space and limited demonstrations. This aligns with previous results from \cite{arunachalam2022dexterous}, and non parametric metrics before better. All the base policies have low performance because we evaluated task is different than the one the demonstration was collected on. Therefore, the policies trained through IRL are the one that can adapt to the changes. 

\subsection{How important is guided exploration?}
\label{subsec:controlled_exp}
\begin{figure}[t!]
  \begin{center}
    \includegraphics[width = \linewidth]{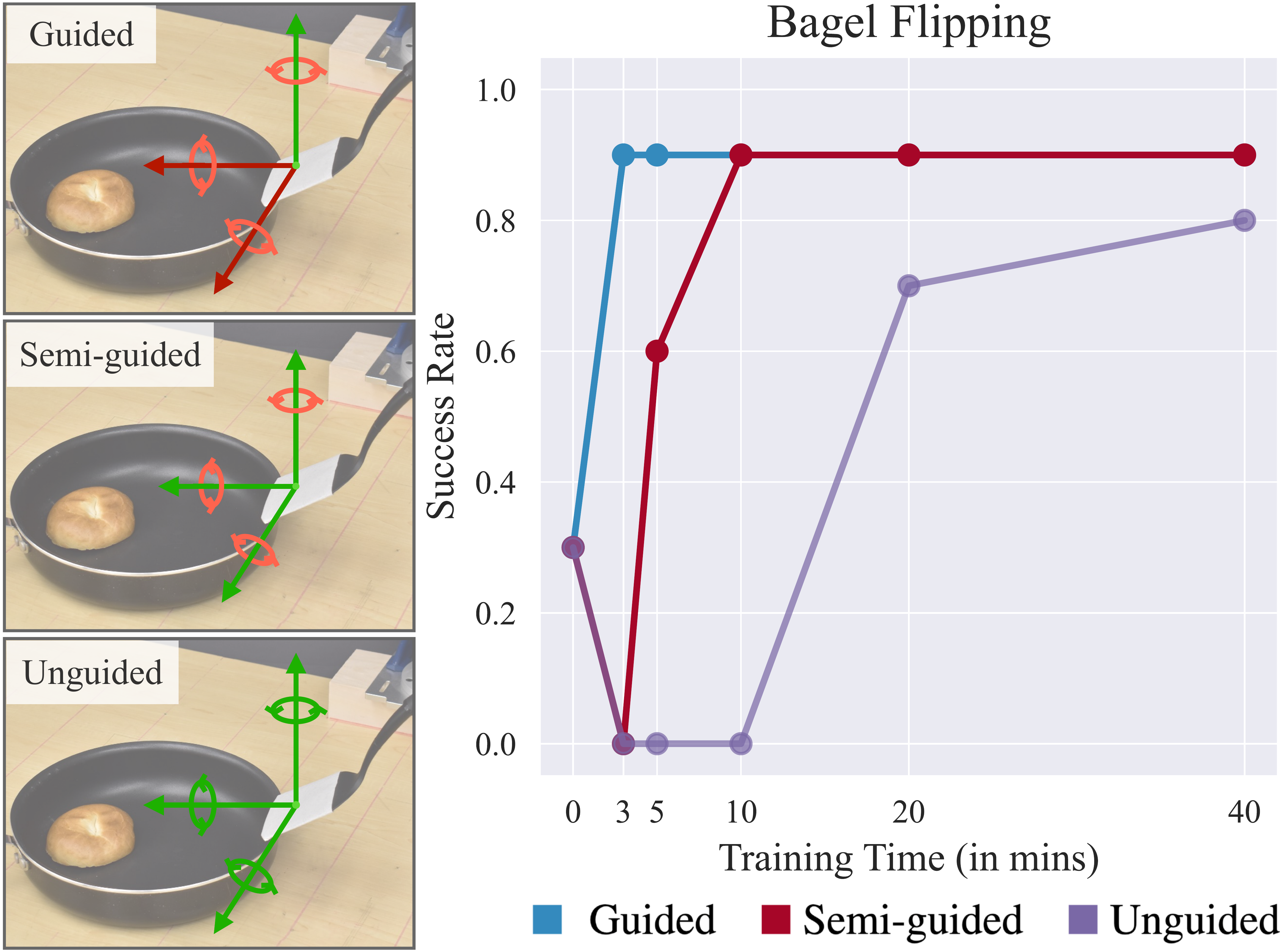}
  \end{center}
  % \begin{center}
  %   \cblock{52}{138}{189}\hspace{1mm} Guided\hspace{1.5mm}
  %   \cblock{166}{6}{40}\hspace{1mm}Semi-guided\hspace{1.5mm}
  %   \cblock{122}{104}{166}\hspace{1mm}Unguided\hspace{1.5mm}
  % \end{center}
  % \cblock{128}{128}{128}\hspace{1mm}Expert\hspace{1.5mm}
  \caption{Comparison between success rate for varied levels of guidance applied to the residual policy. On the left, we show the meaning of each level of guidance. In each scenario, the green axes denote the direction along which the offsets are learned.}
\label{fig:controlled_exploration}
\end{figure}

As opposed to finetuning a parametric model where any update to the model can affect all dimensions of the action space, learning residuals over a fixed-based policy allows us to guide our exploration. For instance, owing to its high dimensional action space, exploring along all dimensions of the action space in the Allegro Hand renders online learning ineffective. So depending on the base policy performance, we only apply residuals along some dimensions while keeping the base policy unaltered along the remaining dimensions. Specifically, we divide our evaluations into three parts - guided, semi-guided, and unguided. For the bagel flipping task, we explore only along the Z-axis for the guided setting, along the XYZ axes for the semi-guided setting, and along both the XYZ axes and roll-pitch-yaw for the unguided setting. Figure~\ref{fig:controlled_exploration} demonstrates the effectiveness of such guided exploration over the unconstrained alternative. Note that although guided exploration improves sample efficiency, unguided exploration with \method{} still outperforms our strongest baselines in Table~\ref{table:base_experiments}.

% Table~\ref{tab:controlled_exploration} provides the dimensions along which residuals were applied on selected tasks and Figure~\ref{fig:controlled_exploration} demonstrates the effectiveness of such controlled exploration. 

% \begin{table}[]
% \caption{\label{tab:controlled_exploration} Table for axes of controlled exploration}
% \begin{tabular}{@{}cccc@{}}
% \toprule
% Task         & \begin{tabular}[c]{@{}c@{}}Controlled \\ Exploration\end{tabular} & \begin{tabular}[c]{@{}c@{}}Semi-Controlled\\ Exploration\end{tabular} & \begin{tabular}[c]{@{}c@{}}Uncontrolled\\ Exploration\end{tabular}                \\ \midrule
% Bagel Flipping   & Only Z-axis                                                       & Along XYZ-plane                                                       & \begin{tabular}[c]{@{}c@{}}Both XYZ-plane and \\ roll-pitch-yaw axes\end{tabular} \\
% Dollar Bill\\ Sliding & Only Z-axis                                                       &   XYZ-thumb                                                                  & XYZ-All                                                                             \\ \bottomrule
% \end{tabular}
% \end{table}

% In the presence of domain knowldege of the task our method allows the user to confine the exploration to only a certain subspace of the action space. This allows for safter exploration, and efficenty in the number of interactions required. In \ref{table:controlled_exp} we demonstrate that exploring in the full action space is inefficent. 

% \begin{figure*}[th]
%   \begin{center}
%     \includegraphics[width = \linewidth]{figures/ot_figure.pdf}
%   \end{center}
%   \caption{An analysis of the values of OT rewards for different trajectories with respected to a given expert demonstration. The leftmost column depicts the visual demonstration, while the other columns each depict a trajectory rollout. Trajectories are sorted in increasing order of OT rewards from left to right. Raw OT scores can be visualized using the red-to-green color map.}
% \label{figure:ot_reward}
% \end{figure*}

\subsection{Is adaptive regularization beneficial for learning offsets?}
In Figure~\ref{fig:controlled_exploration}, we observe that though the base policy has non-zero performance in the bagel flipping task, applying offsets drives the performance to zero during the initial part of online learning. This is primarily due to the untrained offsets driving the agent to an observation unseen in the expert demonstration, thus, adversely affecting the base policy. Drawing inspiration from recent work that uses adaptive regularization to keep the online policy close to the base policy during the initial part of training~\cite{haldar2022watch}, we adaptively regularize our residuals to stay close to zero using the same soft Q-filtering approach (see more details in Appendix~\ref{appendix:soft_qfilter}). However, as observed in Table~\ref{table:bc_reg}, this harms the performance of our model. Empirically, we observe that such regularization drives the residuals to be a very small value close to zero which renders them ineffective in producing significant performance gains over the base policy.

\subsection{How does the choice of base policy affect performance?}
\label{subsec:choice_of_base_policy}
To understand the effect of using different base policies, we compare the performance of \method{} on four variants shown in Table~\ref{table:choice_of_base_policy}. The finetuned ImageNet~\cite{deng2009imagenet} encoder refers to a pretrained ImageNet encoder finetuned with BYOL~\cite{grill2020bootstrap} on the expert demonstrations. These experiments provide 3 key insights - $(a)$ OT-based IRL without pre-training does not work well with few environment interactions, $(b)$ self-supervised learning (SSL) methods such as BYOL do not work well in the low data regime, and $(c)$ with a decent BC policy as in the case of bagel flipping, \method{} can produce significant improvements on the base policy. However, using a non-parametric base policy such as VINN obtains a superior performance as compared to parametric alternatives.

\begin{table}[t!]
\centering
\caption{\label{table:choice_of_base_policy} Comparison between success rates on 10 trials for our method with different base polices.}
\begin{tabular}{c|cc}
\hline
Method          & \begin{tabular}[c]{@{}c@{}}Bagel \\ Flipping\end{tabular} & \begin{tabular}[c]{@{}c@{}}Dollar Bill\\ Picking\end{tabular} \\ \hline
IRL Scratch     & 0.0                                                        & 0.0                                                            \\
Open-loop & 0.1                                                        & \textbf{0.8}  \\
BC              & 0.7                                                       & 0.0                                                            \\
VINN (ImageNet)   & 0.0                                                        & 0.0                                                            \\
VINN (BYOL)        & 0.0                                                        & 0.0                                                            \\
VINN (BC Encoder)     & \textbf{0.9}                                                       & 0.0\\ \hline

\end{tabular}
\end{table}

\subsection{Are pretrained encoders useful for online learning?}
We compare the performance of \method{} with the VINN base policy obtained from a variety of off-the-shelf encoders pretrained using self-supervised learning on large-scale datasets - ImageNet~\cite{deng2009imagenet}, MVP~\cite{Xiao2022, Radosavovic2022} and R3M~\cite{nair2022r3m}. Table~\ref{tab:pretrained} shows that even though these encoders are trained on large-scale datasets, they do not perform well in this setting. In many cases, the performance is worse than our base policies. This is perhaps because the representations learned on Internet data may not transfer well to our suite of tasks. Further, this indicates that representations trained on in-domain data, even in the low-data regime, may perform better than training on large amounts of out-of-domain data. 

% In addition, while BC as a base policy performs poorly, as shown in Table \ref{table:base_experiments}, the representations it learns provide a sufficient signal of task completion as shown in Figure \ref{figure:ot_reward}. We notice that BC completes the coarse portions of the task like going towards the object, but for precise actions it often is inaccurate. Therefore, the representations it learns  may not be detailed enough to perform well in moments of high precision, yet the representations still sufficient for our model. 

% While self supervised represenation learngng (SSL) is a power method to train encoders, they often require ample amounts of data \cite{}. Prior works like VINN \cite{pari2021surprising} demonstrate the effectiness of SSL methods for imitation learning. Because only use between 1 to 3 demos, which amounts to less than 100 observations, this may not be enough to learn a good feature space. However, a more task oriented form of training like BC yields better results as shown in \ref{table:ssl}. Additionally the tasks we solve may not be in the same distribution as the dataset the pretrained model were trained on. 

\begin{table}[t!]
\centering
\caption{\label{tab:pretrained} Analysis of the performance of \method{} using different pre-trained encoders.}
\begin{tabular}{@{}ccc@{}}
\toprule
Encoder  & Bagel Flipping & \begin{tabular}[c]{@{}c@{}}Dollar Bill\\ Picking\end{tabular} \\ \midrule
ImageNet & 0.0             &   0.0                                                          \\
R3M      & 0.0            &   0.1                                                            \\
MVP      & 0.3            &   0.0                                                           \\
BC       & \textbf{0.9}            &   \textbf{0.8}                                                           \\ \bottomrule
\end{tabular}
\end{table}

\subsection{How do additional implementation details affect \method{}?}
\label{subsec:insights}
Table~\ref{tab:insights} provides additional insights with regard to - $(a)$ having a fixed encoder during online learning, $(b)$ conditioning the residual policy on the base policy action. We observe that both of these techniques are necessary and dropping either of them adversely affects the performance of the algorithm.

\begin{table}[t!]
\centering
\caption{\label{tab:insights} Ablation analysis on fixing encoders and conditioning on base actions during online learning.}
\begin{tabular}{@{}cccc@{}}
\toprule
Fix Encoder & \begin{tabular}[c]{@{}c@{}}Condition on\\ base action\end{tabular} & Bagel Flipping & Dollar Bill Picking \\ \midrule
\checkmark         & $\times$                                                                 & 0.6            &  0.1                   \\
$\times$          & \checkmark                                                                &  \textbf{0.9}              &  0.0                   \\
\checkmark         & \checkmark                                                                & \textbf{0.9}            & \textbf{0.8}                 \\ \bottomrule
\end{tabular}
\end{table}

% We demonstrate the efficacy of \method{} in generalizing to varied object configurations unseen in the expert demonstrations. In Figure~\ref{fig:variations}, we show this generalization for a representative task on the xArm and the Allegro Hand. Even though the expert demonstrations only sparsely cover the operating region, we observe to \method{} generalizes to a majority of the designated area of operation.

\begin{figure*}[t]
  \begin{center}
    \includegraphics[width = 1.0\linewidth]{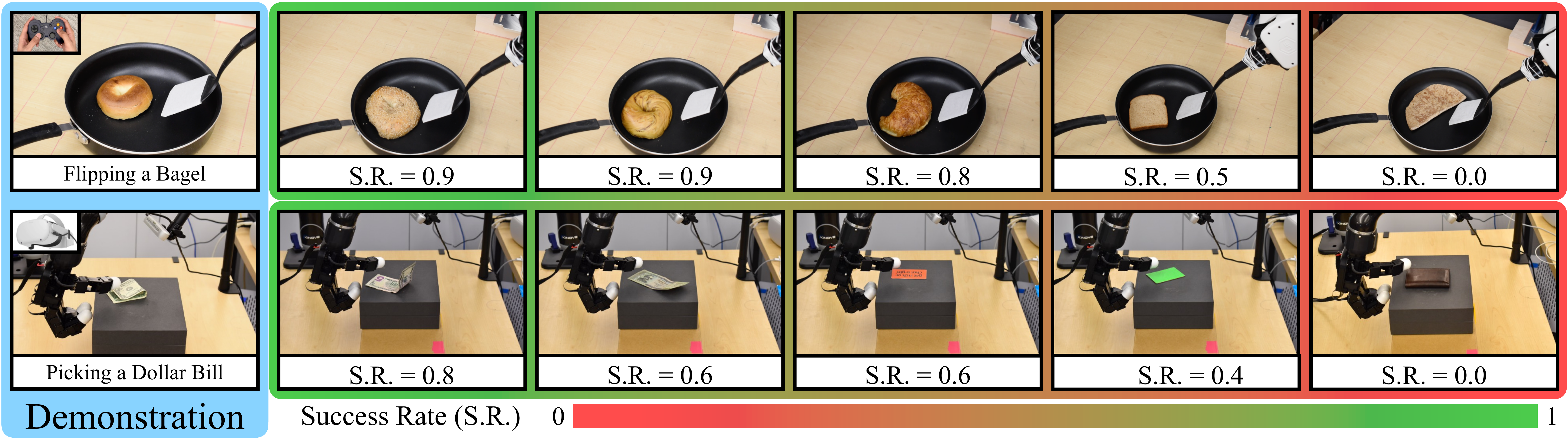}
  \end{center}
  \caption{Here we run \method{}, showing one demonstration on the leftmost object and then training it on new objects. Success rates (S.R.) after 5 minutes of online learning are reported below the corresponding object.}
\label{fig:variations}
\end{figure*}

\subsection{Does \method{} generalize to new objects?}
\label{subsec:variations}
We demonstrate the ability of \method{} to generalize to different objects with varied appearances and dynamics. In Figure~\ref{fig:variations}, we show this generalization for a representative task on the xArm and the Allegro Hand. We observe that the performance drops proportionally with the increase in variation. For instance, the xArm completely fails at flipping a flatbread which is considerably softer than a bagel and requires a different strategy to flip. Similarly, the hand fails to pick up a wallet that is thicker and more uneven than a dollar bill. However, even though the model fails in extreme cases, it succeeds at performing the task with a significant variation in visual and dynamic properties of the object.

% Figure~\ref{fig:variations} shows that given a set of expert demonstrations, \method{} can be trained on different objects with varying sizes and visual features. We observe that this kind of generalization is more effective with small changes in features while the performance drops for extreme changes. For instance, using a soft flatbread with the bagel picking demo does not get any success because the significant change in object dynamics renders the current expert strategy ineffective. A similar observation is made while picking up a wallet while using the demonstration of picking a dollar bill.

\subsection{Limitations of \method{}}
\label{sec:limitations}
To summarize our experiments, we showcase the effectiveness of our algorithm when operating in a low data regime with a limited budget for environment interactions. We demonstrate a significant improvement in performance as compared to prior state-of-the-art work and provide extensive ablations to justify our design choices. However, we recognize a few limitations in this work: $(a)$ Since the OT-based rewards used to train the residual policy align the agent with the demonstrations, it relies on the demonstrator being an ‘expert’. $(b)$ We restrict ourselves to the visual domain which makes it difficult to perform precise tasks where the visual signals are not very prominent. For example, it is difficult to infer a keyhole spanning a minuscule portion of an image. A potential improvement along this line might result from embracing other modalities such as tactile sensing. $(c)$ Our residual policy is randomly initialized. Pretraining the residual policy might help scale to more difficult tasks requiring more precise control.

\section{Related Work}
\textbf{Imitation Learning (IL)}
IL~\cite{argall2009survey, hussein2017imitation} has been shown to solve complex tasks in real-world environments. Approaches for IL include Behavior Cloning (BC)~\cite{pomerleau1998autonomous,torabi2019recent} and Inverse Reinforcement Learning (IRL)~\cite{ng2000algorithms,abbeel2004apprenticeship}. 
BC solely learns from offline demonstrations and has shown promising results in the presence of large diverse datasets~\cite{pomerleau1998autonomous, torabi2018behavioral, codevilla2019exploring, sasaki2021behavioral, young2022playful}. Assistive tools and other teleoperation methods have allowed for more efficient data collection \cite{young2020visual, arunachalam2022holo}. BC has also been applied to tasks with a multimodal action distribution~\cite{florence2022implicit, shafiullah2022behavior, cui2022play}. However, BC suffers on out-of-distribution samples~\cite{ross2011reduction} which renders it unsuitable for the low-data regime. Lately, there has been some work on utilizing non-parametric models to tackle such low data regimes with offline imitation~\cite{pari2021surprising, arunachalam2022dexterous, arunachalam2022holo}. A significant drawback of offline IL is that they do not provide any means for correcting the behavior on unseen observations. IRL provides a solution to this problem by learning a robust reward function through online interactions but suffers from sample inefficiency~\cite{kostrikov2018discriminator}. There has been some work on improving the sample efficiency of IRL~\cite{kostrikov2018discriminator, fu2017learning,xiao2019wasserstein,torabi2018generative}, with some visual extensions to these IRL approaches~\cite{haldar2022watch, cetin2021domain,toyer2020magical,rafailov2021visual,cohen2022imitation}. There has also been demonstration of such IRL approaches performing complex tasks on real robots~\cite{abbeel2004apprenticeship, haldar2022watch, kumar2022graph}.

% Behavior cloning has showed promising results in the presence of large diverse datasets. Works such as \cite{cui2022play, shafiullah2022behavior, young2022playful} show the robustness of behavoir cloning on complex task including multimodal ones. On low data setting there has been recent work on utilizing robust non parametric models instead of parametric ones, works such as \cite{arunachalam2022dexterous, arunachalam2022holo} demonstrate the effectiveness of simple nearest neighbor based models on complex tasks. A significant drawback of offline imitation learning methods is there is no means to correct their behavior. To address this issue, inverse reinforment learning (IRL) allows for online interactions to match the expert behavoir. Works such as \cite{abbeel2004apprenticeship, kumar2022graph, haldar2022watch} show the capabilities of IRL on real world robots. In addition, there have been many works that aim to improve imitation learning through representation learning. Works such as \cite{pari2021surprising, young2022playful} show that representation learning on both task specific data and task agnostic data can improve representations that are used by the policy.

\textbf{Optimal Transport (OT)}
OT~\cite{villani2009optimal,peyre2019computational} provides a tool for comparing probability measures while including the geometry of the space. In imitation learning, OT can be used to compute the alignment between a set of agent and expert observations using distance metrics such as Sinkhorn~\cite{cuturi2013sinkhorn}, Gromov-Wasserstein~\cite{peyre2016gromov}, GDTW~\cite{cohen2021aligning}, CO-OT~\cite{redko2020co} and Soft-DTW~\cite{cuturi2017soft}. Many of these distance metrics have an associated IL algorithm - SIL~\cite{papagiannis2020imitation} uses Sinkhorn, PWIL~\cite{dadashi2020primal} uses greedy Wasserstein, GDTW-IL~\cite{cohen2021aligning} uses GDTW, and GWIL~\cite{fickinger2021cross} using Gromov-Wasserstein. Recent work by ~\citet{cohen2022imitation} has demonstrated that the Sinkhorn distance~\cite{papagiannis2020imitation} produces the most efficient learning among the discussed metrics and can be combined with offline pretraining to efficient perform complex tasks in the real world~\cite{haldar2022watch}. OT has also seen use in the field of computer vision~\cite{sanjabi2018convergence, arjovsky2017wasserstein} to show improvements for Generative Adversarial Networks (GANs) \cite{goodfellow2020generative}. In this work, we adopt the Sinkhorn metric for online learning and combine it with non-parametric IL approaches to perform precise tasks across three robot morphologies.

% Optimal Transport proved to be a useful technique to compute distances on both discrete and continuous distributions. Previous methods included Kullback Leibler divergence and Jensen–Shannon divergence which require the distributions to overlap. However, optimal tranport does not have that constraint. Consequently, it has found useful applications fields such as computer vision (CV) which previously relied on the aforementioned methods for computing distances. Works such as \cite{sanjabi2018convergence, arjovsky2017wasserstein} show improvements for Generatetive Adversial Networks (GANs) \cite{goodfellow2020generative} when incorporating optimal transport techniques to compute distances. In addition, work such as \cite{cohen2021imitation, cohen2022imitation} show optimal transport techniques being used to compute trajectory similarity for imitation learning. For computational speedup often an approximation to the optimal transport is made in the form of entropic regularization which can be solved quickly by algorithms such as \cite{cuturi2013sinkhorn} due to its convexity. 

\textbf{Residual RL for robotics} Learning residuals through RL enables safe and robust online learning~\cite{silver2018residual, johannink2019residual, zhang2019deep, alakuijala2021residual}. Residual RL operates by applying offsets on top of a base policy. Prior works either use a hand-engineered controller~\cite{silver2018residual, johannink2019residual} or a policy learned from demonstrations~\cite{alakuijala2021residual} as the base policy. In this work, we resort to the latter and use non-parametric base policies obtained from one minute of expert demonstration. Prior works also assume the availability of task-specific rewards for learning the online policy. However, we differ from this and use OT matching to obtain rewards from the collected demonstration set.

% Works such as \cite{silver2018residual, johannink2019residual, zhang2019deep, alakuijala2021residual} use various robust base polices and learn offsets to improve the performance of the base policy. These works assume there are environment rewards which in the real world often requires hand crafted functions. Some of these works utilize controllers as the base policy. Unlike works like \cite{haldar2022watch}, these simple yet robust controllers as base policies can not be directly finetuned and require offsets. Residual policies have proven to enable online interaction and exploration in a safe manner by restricting the range of the offsets. 

% \textbf{Adaptation}
% The ability to adapt is a critical feature of any intelligent system. In the field of robotics. One area of research is using meta learning for adaption. Works, such as \cite{kaushik2020fast} use meta learning to adapt a dynamical model in simulation and use this to deploy their agent in the real world. In addition, \cite{yu2020learning} uses meta learning to adapt a latent conditioned policy to new environments and reward functions. \cite{kumar2022adapting} learns an entristics model, and adapts the base policy through model free RL. \cite{qi2022hand} leanrn a controller in simulation to rotate objects and finetunes the controller through proprioception data in the real world on new objects. 

% \section{Discussion and Limitations (Siddhant) (0.25-0.5)}
% \input{documents/discussion.tex}

\section{Conclusion}
\label{conclusion}

In this work, we present a new algorithm for fast imitation learning, FISH, that demonstrates improved performance compared to prior state-of-the-art work on a variety of real robot tasks across three different robot morphologies. We demonstrate that combining an imperfect base policy with a learned residual policy can enable performing precise tasks with one minute of demonstration collection and limited environment interactions. Further, we ablate over various design decisions of \method{}, which shows the importance of learning stable representations, choosing the right base policy, and performing guided exploration. While powerful, we recognize that \method{} has limitations (see Section~\ref{sec:limitations}).
% We believe that further research into developing better visual representations, perhaps through large robot models, could improve generalization across object categories along with object configurations.

% A potential future direction for our work is combining such a residual framework with bulky offline models performing complex tasks with large amounts of data. Another interesting direction would be to extend this line of research to generalize across naturally occurring variations in tasks such as the color or shape of an object and changing backgrounds, among  others. In addition, changing the dynamics of the objects and environment would be another interesting axis to generalize on.

\section*{Acknowledgments}
We thank Sridhar Pandian Arunachalam, David Brandfonbrener, Zichen Jeff Cui, Venkatesh Pattabiraman, Ilija Radosavovic, and Chris Paxton for valuable feedback and discussions. This work was supported by grants from Honda, Meta, Amazon, and ONR awards N00014-21-1-2758 and N00014-22-1-2773.

%% Use plainnat to work nicely with natbib. 

\bibliographystyle{plainnat}
\bibliography{references}

\newpage
\appendix

\subsection{Background}

\subsubsection{Optimal Transport for Imitation Learning~(OT)}
\label{appendix:optimal}
To alleviate the non-stationary reward problem with adversarial IRL frameworks, a new line of OT-based approaches has been recently proposed~\cite{papagiannis2020imitation,dadashi2020primal,cohen2022imitation}. Intuitively, the closeness between expert trajectories $\mathcal{T}^e$ and behavior trajectories $\mathcal{T}^b$ can be computed by measuring the optimal transport of probability mass from $\mathcal{T}^b \rightarrow \mathcal{T}^e$. During policy learning, an encoder $f_{\phi}$ transforms observations into informative state representations. Some examples of a preprocessor function $f_{\phi}$ are an identity function, a mean-variance scaling function and a parametric neural network. In this work, we use a parametric neural network as $f_{\phi}$. Given a cost function $c:\mathcal{O}\times\mathcal{O}\rightarrow\mathbb{R}$ defined in the preprocessor's output space and an OT objective $g$, the optimal alignment between an expert trajectory $\textbf{o}^{e}$ and a behavior trajectory $\textbf{o}^{b}$  can be computed as 

\begin{equation}
    \mu^{*} \in \underset{\mu\in\mathcal{M}}{\text{arg~min}}~g(\mu, f_{\phi}(\textbf{o}^{b}), f_{\phi}(\textbf{o}^{e}), c)
    \label{appendix:eq:alignment}
\end{equation}

where $\mathcal{M}=\{\mu\in\mathbb{R}^{T\times T}:\mu\boldsymbol{1}=\mu^{T}\boldsymbol{1}=\frac{1}{T}\boldsymbol{1}\}$ is the set of coupling matrices and the cost $c$ can be the Euclidean or Cosine distance. The optimal alignment $\mu^{*}$ is subjected to the constraints  $\sum_i\mu^*_{ij} = P(\textbf{o}^e_i)~\forall i$ and $\sum_j\mu^*_{ij} = P(\textbf{o}^b_j)~\forall j$. In this work, inspired by \cite{cohen2022imitation}, we use the entropic Wasserstein distance~\cite{cuturi2013sinkhorn} with cosine cost as our OT metric, which is given by the equation

\begin{equation}
\begin{aligned}
    g(\mu, f_{\phi}(\textbf{o}^{b}), f_{\phi}(\textbf{o}^{e}), c) &=  \mathcal{W}^{2}(f_{\phi}(\textbf{o}^{b}), f_{\phi}(\textbf{o}^{e}))\\
    &= \sum_{t,t'=1}^{T}C_{t,t^{'}} \mu_{t,t'}
\end{aligned}
\label{appendix:eq:wasserstein}
\end{equation}

where the cost matrix $C_{t,t^{'}} = c(f_{\phi}(\textbf{o}^{b}), f_{\phi}(\textbf{o}^{e}))$. Using Eq.~\ref{appendix:eq:wasserstein} and the optimal alignment $\mu^{*}$ obtained by optimizing Eq.~\ref{appendix:eq:alignment}, a reward signal can be computed for each observation using the equation
\begin{equation}
    \label{appendix:eq:ot_reward}
    r^{OT}(o^{\boldsymbol{b}}_{t}) = - \sum_{t'=1}^{T} C_{t,t^{'}} \mu^{*}_{t,t^{'}}
\end{equation}

Intuitively, maximizing this reward encourages the imitating agent to produce trajectories that closely match demonstrated trajectories. Since solving Eq.~\ref{appendix:eq:alignment} is computationally expensive, approximate solutions such as the Sinkhorn algorithm~\cite{knight2008sinkhorn,papagiannis2020imitation} are used instead.

\subsubsection{Soft Q-filtering}
\label{appendix:soft_qfilter}
In ROT~\cite{haldar2022watch}, a soft Q-filtering based approach is used for appropriately weighing the Q-value based actor-critic loss and the BC regularization loss in the objective function. The key idea behind this is to keep the online learning policy $\pi^{ROT}$ close to the behavior policy on regions of the observation space where the behavior policy exhibits better performance than $\pi^{ROT}$. More precisely, given a behavior policy $\pi^{BC}(s)$, the current policy $\pi^{ROT}(s)$, the Q-function $Q(s,a)$ and the replay buffer $\mathcal{D}_{\beta}$, the BC regularization weight $\lambda$ is computed as:

\begin{equation}
    \lambda(\pi^{ROT}) = \mathbb{E}_{(s,\cdot)\sim \mathcal{D}_{\beta}}\left[\mathbbm{1}_{Q(s,\pi^{BC}(s))>Q(s,\pi^{ROT}(s))} \right]
\end{equation}

 This filtering strategy is inspired by \citet{nair2018overcoming}, which employs a binary hard assignment instead of a soft continuous weight. In Table~\ref{table:bc_reg}, we provide an ablation study comparing the performance of \method{} with and without such a regularization scheme. In \method{}, since we learn a residual policy on top of a fixed base policy during online learning, such a regularization scheme must keep the final action, which is the sum of the base action and the residual offset, close to the action in the expert demonstrations. To do this, we apply the regularization such that the offsets are regularized to stay close to zero when the base policy $\pi^b$ is performing better than the policy $\pi^{\method{}}$. Table~\ref{table:bc_reg} shows that applying such offset regularization harms the performance of our model. Empirically, we observe that such regularization drives the residuals to be a very small value close to zero which renders them ineffective in producing significant performance gains over the base policy.

%%%%%%%%%%%%%%%%%%%%%%%%%%%%%%%%%%%%%%%%%%%%%%%%%%%%%%%%%%%%%%%%%%%%%%%%%%%%%%%%%%%%%%%%%%%%%%%%%%%%

%%%%%%%%%%%%%%%%%%%%%%%%%%%%%%%%%%%%%%%%%%%%%%%%%%%%%%%%%%%%%%%%%%%%%%%%%%%%%%%%%%%%%%%%%%%%%%%%%%%%

\subsection{Safe exploration using a guided residual policy}
\label{appendix:sec:safe_exploration}
\begin{figure}[t]
  \begin{center}
    \includegraphics[width = \linewidth]{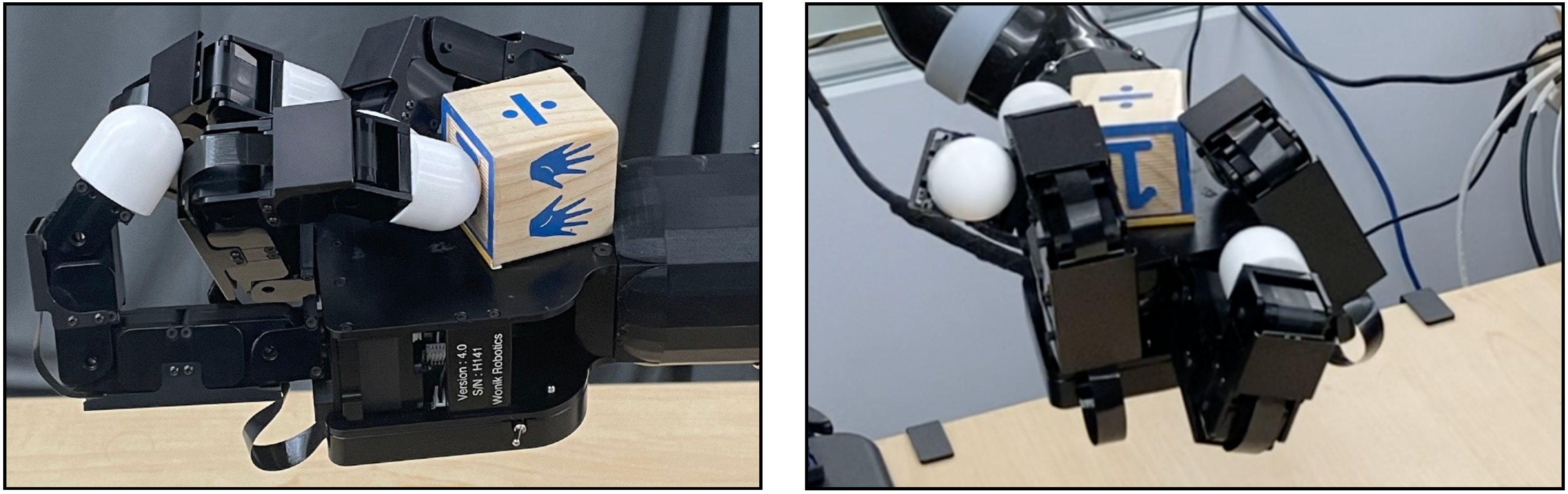}
  \end{center}
  \caption{Snapshots of the Allegro Hand while training an online policy using ROT. Here, we show that exploring along all action dimensions results in unsafe learning on the hand, driving it to undesirable positions and unnatural poses.}
\label{appendix:fig:safety}
\end{figure}

In \method{}, we propose guiding the residual policy to learn offsets along a smaller subspace of the full action space. In other words, the residual policy is used to learn offsets only along certain action dimensions instead of the complete action space. In addition to providing performance benefits, as shown in Figure~\ref{fig:controlled_exploration}, we argue that such guided exploration also enables safer online learning. For instance, Figure~\ref{appendix:fig:safety} shows the Allegro hand during different stages of online learning while training ROT. ROT explores along all dimensions of the action space which drives to hand to undesirable positions, resulting in collisions between fingers and unnatural poses. This confirms that guided exploration of the action space, as proposed in \method{}, enables safer online training of such systems.

%%%%%%%%%%%%%%%%%%%%%%%%%%%%%%%%%%%%%%%%%%%%%%%%%%%%%%%%%%%%%%%%%%%%%%%%%%%%%%%%%%%%%%%%%%

\subsection{Algorithm Details}
\label{appendix:algo_details}

\subsubsection{Algorithm block}
\label{appendix:sec:algorithm_block}
Algorithm~\ref{alg:fish} describes our proposed algorithm, Fast Imitation of Skills from Humans (FISH).

\begin{algorithm}[ht!]
\caption{FISH: Fast Imitation of Skills from Humans}\label{alg:fish}
\begin{algorithmic}
    \State $\textbf{Require:}$\\
    Expert Demonstrations $\mathcal{T}^e \equiv \{(o_t, a_t)_{t=0}^{T}\}_{n=0}^N$\\
    BC pretrained Encoder $f_{enc}$\\
    Base policy $\pi^{b}$\\
    Replay buffer $\mathcal{D}$, Training steps $T$, Episode Length $L$\\
    Task environment $env$\\
    Parametric networks for RL backbone (e.g., the policy and critic function for DrQ-v2)\\

    \State $\textbf{Algorithm:}$
    % \State $\pi^{ROT} \gets \pi^{BC}$ \Comment{Initialize with pretrained policy}
    \State Randomly initialize residual policy $\pi^{r}$  and critic $Q$
    \For{each timestep $t = 1...T$}
        \If{done}
            \State $r_{1:L} = \text{rewarder}_{OT}\text{(episode)}$ \Comment{OT reward}
            \State $\text{Update episode with}~r_{1:L}$
            \State $\text{Add}~(\textbf{o}_{t}, \textbf{a}_{t}, \textbf{a}^{b}_{t}, \textbf{o}_{t+1}, r_{t})~\text{to}~\mathcal{D}$
            \State $\textbf{o}_{t} = env.reset(),~\text{done} = \text{False},~\text{episode} = [~]$ 
        \EndIf
        \State $\textbf{z}_{t} = f_{enc}(\textbf{o}_{t})$
        \State $\textbf{a}^{b}_{t} \sim \pi^{b}(\textbf{z}_{t})$ \Comment{Compute base action}
        \State $\textbf{a}^{r}_{t} \sim \pi^{r}(\textbf{z}_{t}, \textbf{a}^{b}_{t})$ \Comment{Compute residual offset}
        
        \State $\textbf{a}_{t}$ = $~\textbf{a}^{b}_{t} + \textbf{a}^{r}_{t}$
        \State $\textbf{o}_{t+1},~\text{done} = env.step(\textbf{a}_{t})$
        \State $\text{episode.append}([\textbf{o}_{t}, \textbf{a}_{t}, \textbf{a}^{b}_{t}, \textbf{o}_{t+1}])$
        \State $\text{Update backbone-specific networks using}~\mathcal{D}$
    \EndFor
\end{algorithmic}
\end{algorithm}

\subsubsection{Algorithm and training procedure} Our model consists of 3 primary neural networks - the encoder, the actor and the critic. During the BC pretraining phase, the encoder and the actor are trained using a mean squared error (MSE) on the expert demonstrations. Next, for finetuning, the pretrained BC encoder is loaded and is used to obtain the encoded observations. The actor and the critic are initialized randomly. The loaded BC encoder is fixed during online learning in order to stationarize the OT rewards which are computed on the encoded representations.

\subsubsection{Actor-critic based reward maximization} We use a recent n-step DDPG proposed by \citet{yarats2021mastering} as our residual RL backbone. The deterministic actor is trained using deterministic policy gradients (DPG)~\cite{silver2014deterministic}. The critic is trained using clipped double Q-learning similar to ~\citet{yarats2021mastering} in order to reduce the overestimation bias in the target value. This is done using two Q-functions, $Q_{\theta1}$ and $Q_{\theta2}$. The critic loss for each critic is given by the equation

\begin{equation}
    \mathcal{L}_{\theta_{k}} = \mathbbm{E}_{(z,a^b,a^r)\sim D_{\beta}} \left[(Q_{\theta_{k}}(z,a^b,a^r) - y)^{2} \right] \forall ~k \in \{1,2\}
\end{equation}

where $\mathcal{D}_{\beta}$ is the replay buffer for online rollouts and $y$ is the target value for n-step DDPG given by

\begin{equation}
    y = \sum_{i=0}^{n-1} \gamma^{i}r_{t+i} + \gamma^{n} \underset{k=1,2}{min}Q_{\bar \theta_{k}}(s_{t+n}, a^{b}_{t+n}, a^{r}_{t+n})
\end{equation}

Here, $\gamma$ is the discount factor, $r$ is the reward obtained using OT-based reward computation and $\bar \theta_{1}$, $\bar \theta_{2}$ are the slow-moving weights of target Q-networks. 

\subsubsection{Architecture Details} 
Our encoder takes in an $84\times84$ image as input and produces a $512$-dimensional output. The encoder comprises 4 convolutional layers with a single linear layer. The actor takes in the encoded representation along with the action from the base policy and passes it through 3 linear layers to produce an action. The critic takes in the encoded representation, the action from the base policy, and the residual action and passes it through 3 linear layers to produce the Q-value.

\subsubsection{Hyperparameters}

The complete list of hyperparameters used in the paper has been provided in Table~\ref{tab:hyperparams}. 

\begin{table*}[h!]
    \begin{center}
    \setlength{\tabcolsep}{18pt}
    \renewcommand{\arraystretch}{1.5}
    \begin{tabular}{ c c c } 
        \hline
        Method & Parameter & Value \\
        \hline
        Common & Replay buffer size & 5000 \\
               & Learning rate      & $1e^{-4}$\\
               & Discount $\gamma$   & 0.99\\
               & $n$-step returns   & 3\\
               & Action repeat      & 1\\
               & Seed frames        & 260 (xArm, Stretch), 200 (Allegro Hand) \\
               & Mini-batch size    & 256\\
               & Agent update frequency & 2\\
               & Critic soft-update rate & 0.01\\
               & Feature dim        & 50\\
               & Hidden dim         & 1024\\
               & Optimizer          & Adam\\
        \hline
        FISH    & Exploration steps   & 0\\
               & DDPG exploration schedule & 0.1\\
               & Reward scale factor & 10\\
        \hline
        ROT    & Exploration steps   & 0\\
               & DDPG exploration schedule & 0.1\\
               & Reward scale factor & 10\\
               & Fixed weight $\alpha$       & 0.03\\
        \hline
        DAC    & Exploration steps   & 2000\\
               & DDPG exploration schedule & linear(1,0.1,2000)\\
               & Gradient penalty coefficient & 10\\
        \hline
    \end{tabular}
    \end{center}
    \caption{List of hyperparameters.}
    \label{tab:hyperparams}
\end{table*}

\subsection{Task description}
\label{appendix:sec:task_description}
In this section, we describe the suite of manipulation experiments carried out on a real robot in this paper.

\begin{enumerate}[label=(\alph*),leftmargin=*]
    \item \textbf{Door opening:} The Stretch robot is supposed to open a closed door in a cupboard.
    \item \textbf{Drawer opening:} The Stretch robot is supposed to open a closed drawer in a kitchen. 
    \item \textbf{Light switching:} The Stretch Robot is tasked with switching off a light switch that is in an ON position.
    \item \textbf{Cube flipping:} The Allegro Hand is tasked with flipping a cube placed on its palm by 90 degrees.
    \item \textbf{Bottle cap spinning:} The Allegro Hand is tasked with spinning a bottle cap by 270 degrees.
    \item \textbf{Dollar bill picking:} The Allegro Hand is tasked with sliding and picking up a bill placed on a platform.
    \item \textbf{Peg in a cup:} The xArm is tasked with placing a peg inside a cup which is moved to random positions on a table.
    \item \textbf{Bagel flipping:} The xArm is tasked with using a spatula to flip a bagel placed at random positions on a fry pan.
    \item \textbf{Key insertion:} The xArm is tasked with inserting a key inside the keyhole of a lock that is randomly placed on a table.
\end{enumerate}

\textbf{Evaluation procedure} For each task, we obtained a set of random points where the object is placed (for xArm and the Allegro Hand) or the robot is initialized (for the Stretch robot) and evaluate ~\method{} and all of the other methods on the same set of initializations. These initializations are different for each task based on the limits of the observation space of the task. For the tasks on the xArm and the Allegro Hand, we evaluate on a set of 10 different object positions while the tasks on the Stretch are evaluated on a set of 5 different robot initializations.

\begin{figure*}[h]
  \begin{center}
    \includegraphics[width = \linewidth]{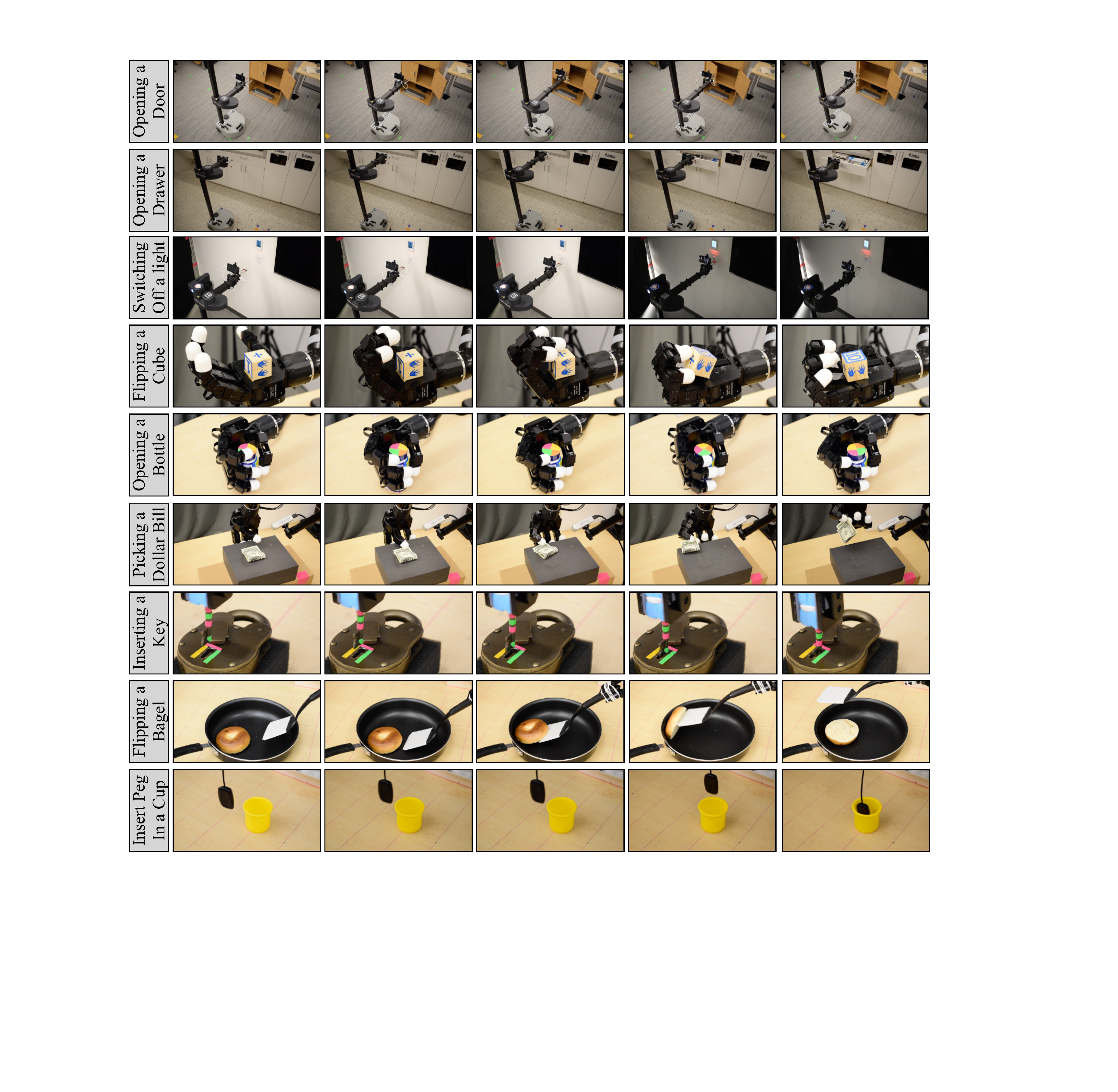}
  \end{center}
  \caption{Robot rollouts for the 9 tasks we evaluate \method{} on.}
\label{appendix:fig:robot_rollout_bagel}
\end{figure*}

\subsection{Generalization to unseen objects}
\label{appendix:sec:unseen_objects}
We evaluate \method{} on a set of unseen objects and demonstrate the generalization capabilities of our method. Figure~\ref{appendix:fig:robot_rollout_bagel} shows the generalization of \method{} to different types of bread despite the demonstration being on a specific kind of bagel. Similarly, Figure~\ref{appendix:fig:robot_rollout_hand} shows such generalization to different bills and cards despite the demonstration set comprising a specific dollar bill.

\begin{figure*}[ht]
  \begin{center}
    \includegraphics[width = \linewidth]{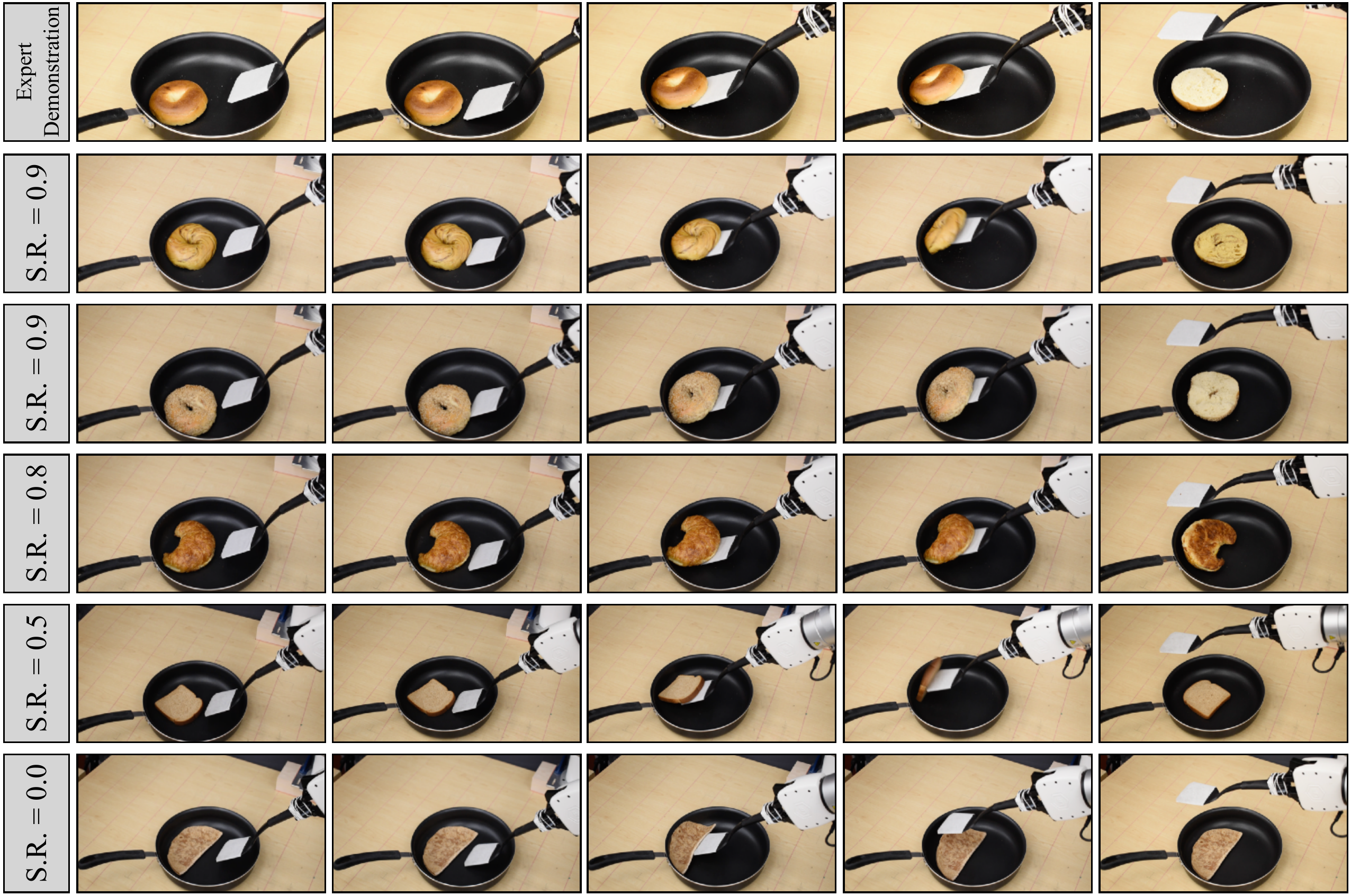}
  \end{center}
  \caption{Generalization of \method{} to different types of bread.}
\label{appendix:fig:robot_rollout_hand}
\end{figure*}

\begin{figure*}[ht]
  \begin{center}
    \includegraphics[width = \linewidth]{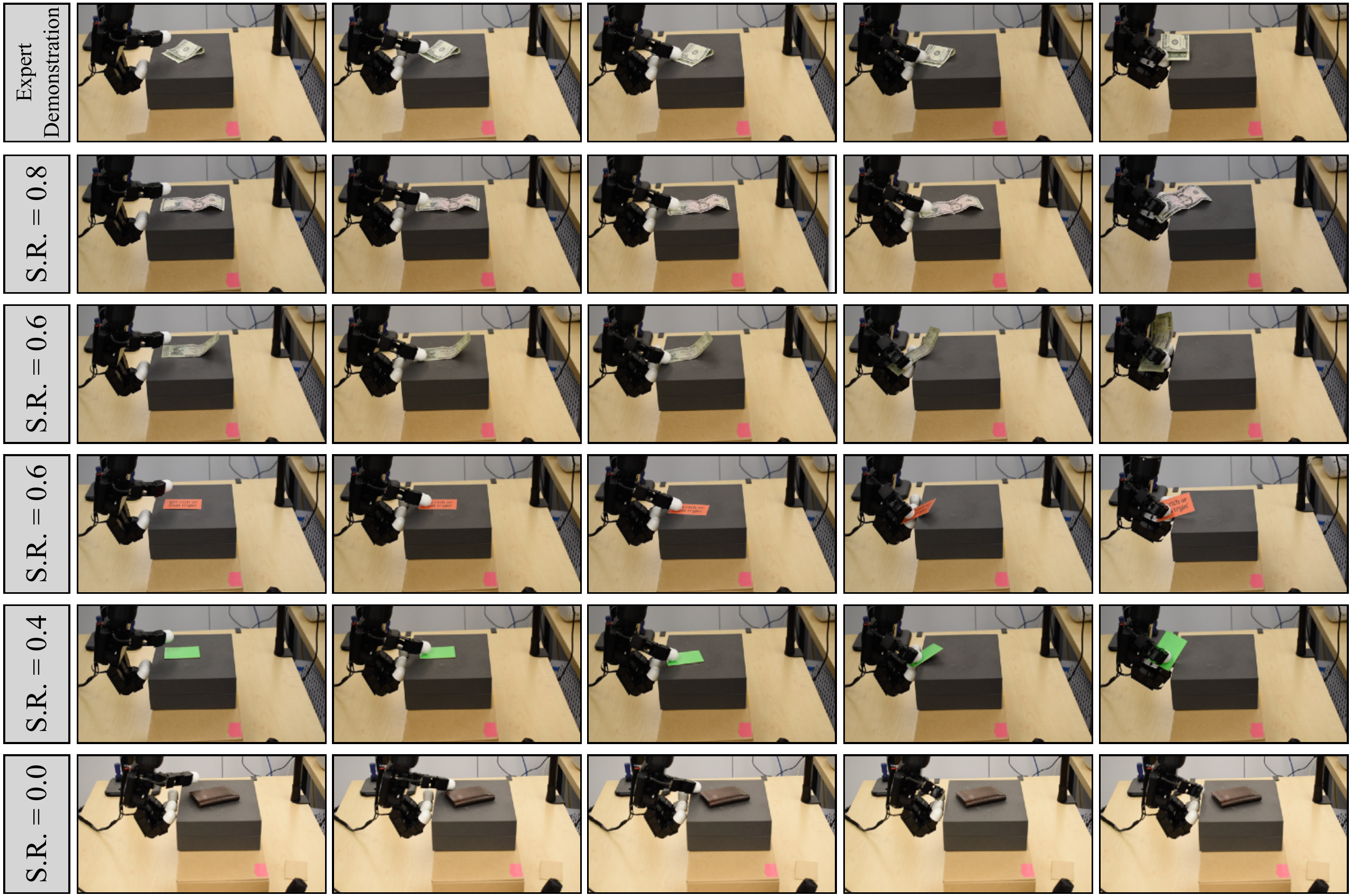}
  \end{center}
  \caption{Generalization of \method{} to different bills and cards.}
\label{appendix:fig:robot_rollout_hand}
\end{figure*}

\end{document}